\documentclass[runningheads]{llncs}

\usepackage{eccv}

\usepackage{eccvabbrv}

\usepackage{graphicx}
\usepackage{booktabs}
\usepackage{iftex}

\ifPDFTeX
  \usepackage[accsupp]{axessibility}
\fi

\usepackage{hyperref}

\usepackage{orcidlink}
\newcommand{\ourmethod}{\textsc{MM-WebAgent}\xspace}
\newcommand{\ourbenchmark}{\textsc{MM-WebGEN-Bench}\xspace}

\usepackage{graphicx}
\usepackage{booktabs}
\usepackage{multirow}
\usepackage[normalem]{ulem}
\useunder{\uline}{\ul}{}
\usepackage{wrapfig}
\usepackage{pifont}
\usepackage{enumitem}
\usepackage{xcolor}
\definecolor{gainblue}{RGB}{33,113,181}

\definecolor{myBlue}{RGB}{0,112,192}

\usepackage[table,xcdraw]{xcolor}
\usepackage{xspace}

\usepackage{authblk}    

\usepackage{listings}
\usepackage{xcolor}
\usepackage{tcolorbox}
\tcbuselibrary{listings,breakable,skins,theorems}
\newcounter{promptctr}
\newtcblisting{promptbox}[2][]{%
  listing engine=listings,
  listing only,
  listing options={
    basicstyle=\ttfamily\footnotesize,
    breaklines=true,
    breakatwhitespace=false,
    columns=fullflexible,
    keepspaces=true,
    showstringspaces=false
  },
  colback=gray!3,
  colframe=gray!40,
  boxrule=0.6pt,
  arc=2mm,
  left=2mm,
  right=2mm,
  top=1mm,
  bottom=1mm,
  breakable,
  enhanced,
  before upper={\refstepcounter{promptctr}},
  title={Prompt~\thepromptctr: #2},
  fonttitle=\bfseries\small,
  #1
}

\newcommand{\std}[1]{\scriptsize\textcolor{gray}{$\pm$#1}}
\usepackage{wrapfig}
\usepackage[T1]{fontenc}
\usepackage[utf8]{inputenc}

\begin{document}

\title{\ourmethod{}: A Hierarchical Multimodal Web Agent for Webpage Generation}

\titlerunning{\ourmethod{}}

\author{
Yan Li\inst{1}$^{*}$ \and
Zezi Zeng\inst{2}$^{*}$ \and
Yifan Yang\inst{4}$^{\dagger}$ \and
Yuqing Yang\inst{4} \and
Ning Liao\inst{1} \and
Weiwei Guo\inst{3} \and
Lili Qiu\inst{4} \and
Mingxi Cheng\inst{4} \and
Qi Dai\inst{4} \and
Zhendong Wang\inst{4} \and
Zhengyuan Yang\inst{4} \and
Xue Yang\inst{1}$^{\dagger}$ \and
Ji Li\inst{4} \and
Lijuan Wang\inst{4} \and
Chong Luo\inst{4}
}

\authorrunning{Y. Li et al.}

\institute{
Shanghai Jiao Tong University
\and
Xi'an Jiaotong University
\and
Tongji University
\and
Microsoft Corporation
}

\maketitle

\begin{center}
\small \url{https://aka.ms/mm-webagent}
\end{center}

\begingroup
\renewcommand\thefootnote{}
\footnotetext{$^{*}$ Equal contribution. This work was done during their internship at Microsoft.}
\footnotetext{$^{\dagger}$ Corresponding to: Yifan Yang <yifanyang@microsoft.com>, Xue Yang <yangxue-2019-sjtu@sjtu.edu.cn>.}
\endgroup

\begin{abstract}
The rapid progress of Artificial Intelligence Generated Content (AIGC) tools enables images, videos, and visualizations to be created \emph{on demand} for webpage design, offering a flexible and increasingly adopted paradigm for modern UI/UX. However, directly integrating such tools into automated webpage generation often leads to style inconsistency and poor global coherence, as elements are generated in isolation. We propose \ourmethod{}, a hierarchical agentic framework for multimodal webpage generation that coordinates AIGC-based element generation through hierarchical planning and iterative self-reflection. \ourmethod{} jointly optimizes global layout, local multimodal content, and their integration, producing coherent and visually consistent webpages. We further introduce \ourbenchmark{} and a multi-level evaluation protocol for systematic assessment. Experiments demonstrate that \ourmethod{} outperforms code-generation and agent-based baselines, especially on multimodal element generation and integration.
  \keywords{Multimodal Webpage Generation \and Agentic Framework \and Generative AI (AIGC)}
\end{abstract}

\section{Introduction}

Webpage generation~\cite{laurenccon2024unlocking,shrivastava2023repository,huang2025seeing,guo2025iw} is a practical and high-impact application of large language models (LLMs): given a natural-language request, modern systems can quickly synthesize HTML/CSS and prototype complete pages. Recent \emph{web agents}~\cite{li2025codetree,zhang2024codeagent,huang2024agentcoder,lu2025webgen,wang2024openhands} further automate this process by decomposing an intent into executable steps. However, real-world webpages are not purely text and code---they contain heterogeneous \emph{multimodal} elements such as images, videos, and charts, whose content, style, and geometry must cohere with the global layout and the semantic intent.

\begin{figure*}[!t]
    \centering
    \includegraphics[width=\linewidth]{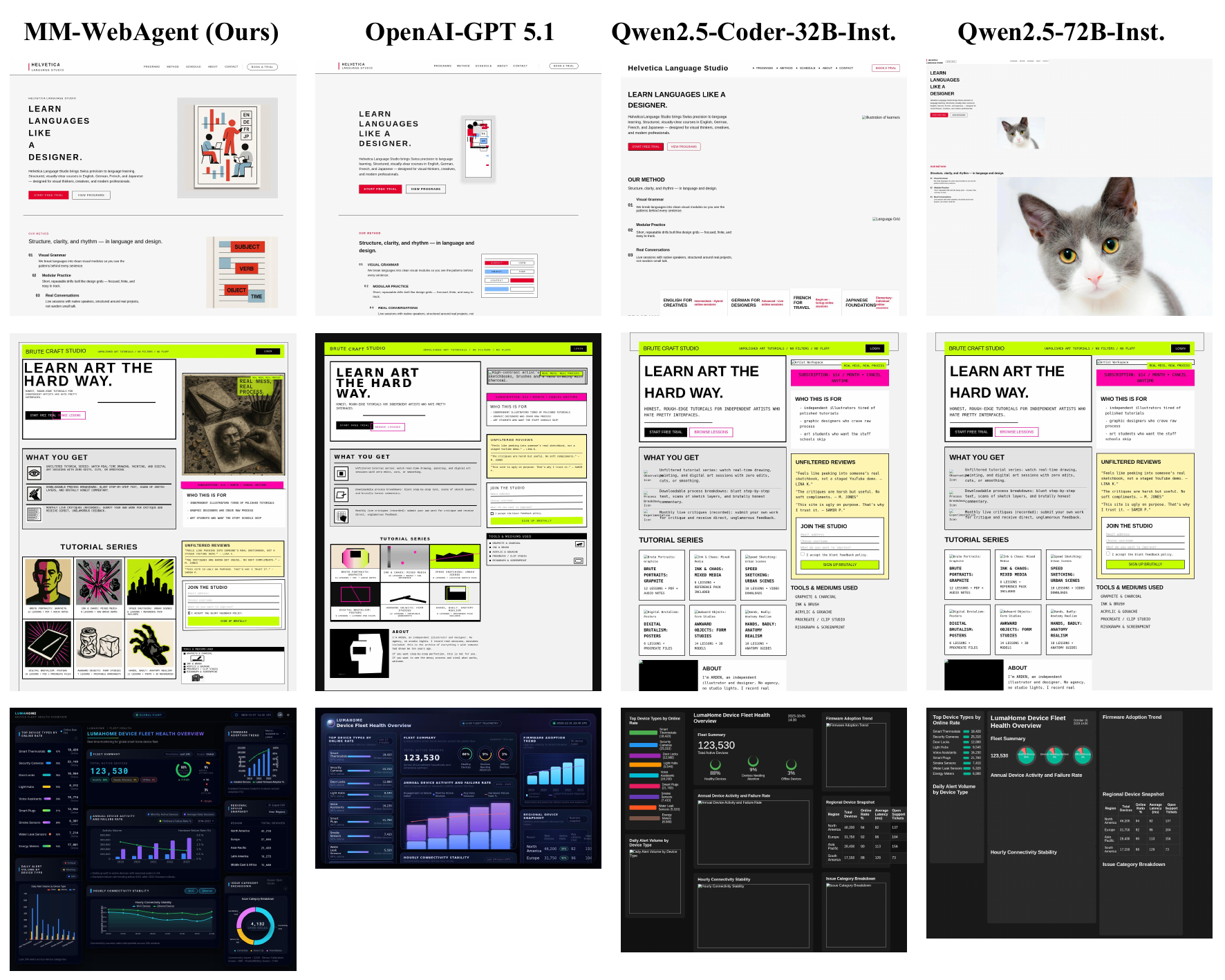}
    \caption{\textbf{Rendered webpage examples generated by \ourmethod{} and baseline methods on \ourbenchmark{}.} \ourmethod{} generates webpages with more coherent layouts, consistent visual styles, and better-integrated multimodal elements compared to baseline methods.}
    \label{fig:vis_comparison}
\end{figure*}

Most existing pipelines populate these elements via retrieval or placeholders, and then generate or insert assets independently. This often leads to (i) \textbf{style inconsistency} across elements, (ii) \textbf{geometry mismatch} between generated media and reserved slots, and (iii) \textbf{global incoherence} after assets are composed into the page. Motivated by the iterative workflow of human designers, we argue that multimodal webpage generation should be treated as a structured \emph{plan-and-refine} process, where global layout decisions and local asset generation are explicitly coordinated and repeatedly refined.

We propose \ourmethod{}, a hierarchical agentic framework that integrates \textbf{hierarchical planning} and \textbf{hierarchical self-reflection} for multimodal webpage generation.
An overview of the framework is shown in Fig.~\ref{fig:framework}.

In the \textbf{hierarchical planning} stage, \ourmethod{} generates a \emph{global layout plan} specifying section hierarchy, ordering, coarse spatial organization, and page-level style attributes, together with placeholders and constraints for multimodal components. Conditioned on this global context, it then constructs \emph{local element plans} for each multimodal component, encoding the element's functional role, surrounding section context, expected size/aspect constraints, and style guidance, enabling downstream generators to produce semantically appropriate and stylistically compatible assets.

To emulate iterative design, \ourmethod{} further performs \textbf{hierarchical self-reflection} at three levels: \emph{(i) local refine} improves individual assets to better satisfy their local plans; \emph{(ii) context refine} patches surrounding HTML/CSS to resolve integration issues (e.g., misalignment, overflow, spacing); and \emph{(iii) global refine} revises the entire page using both the HTML code and rendered screenshots to enhance layout balance and style coherence. This design enables joint optimization of content, geometry, and aesthetics, rather than treating multimodal elements as loosely coupled add-ons.

To support systematic evaluation, we introduce \ourbenchmark{}, a benchmark for multimodal webpage generation spanning diverse intents, layouts, styles, and multimodal compositions.
We further design a \emph{multi-level evaluation protocol} that decomposes webpage quality into \emph{global-level} criteria (layout correctness, style coherence, and aesthetics) and \emph{local-level} criteria for embedded multimodal elements (image, video, and chart quality), enabling fine-grained analysis of both overall page quality and individual components.

Experiments on \ourbenchmark{} show that \ourmethod{} consistently outperforms both code-generation and \emph{code-only agent} baselines, with particularly strong gains on multimodal element generation and integration, highlighting the advantage of enabling agentic coordination with native multimodal asset generation.

Our contributions are three-fold:
(1) we introduce a \emph{multimodal web agent paradigm} that goes beyond code-only generation by enabling hierarchical agentic planning over native multimodal asset generation, coupling global layout planning with context-aware local element planning;
(2) we propose a hierarchical self-reflection mechanism that iteratively refines multimodal webpages at the local, context, and global levels; and
(3) we present \ourbenchmark{} together with a multi-level evaluation protocol for systematic benchmarking of multimodal webpage generation.

\section{Related Work}

\subsection{Visual Code Generation.}
Recent advances in multimodal learning have driven increasing interest in visual code generation for webpages~\cite{xiao2024interaction2code,sun2025fullfront,yun2024web2code}. Existing studies typically incorporate visual information in one of two ways: including reconstructing webpages from screenshots by parsing visual elements into executable HTML code~\cite{huang2025seeing,guo2025iw}, or augmenting webpage generation with externally retrieved visual assets~\cite{openai_gpt51}. While these approaches improve layout fidelity and code correctness, they treat multimodal assets as static or externally provided, limiting their ability to generate novel, semantically aligned, and stylistically coherent multimodal content.

\subsection{Vision-Language Code Agents.}
To manage the complexity of on-demand generation, code agents have been introduced to orchestrate the design process, extending large language models with planning, tool use, and environmental interaction to solve complex tasks~\cite{yang2024swe}.
Recent work such as OpenHands~\cite{wang2024openhands} and Bolt.diy~\cite{stackblitz_bolt_diy} employ hierarchical task planning to decompose software engineering workflows into executable steps, while ReCode~\cite{yu2025recode} unifies planning and action within a single code representation for fine-grained control.
In the context of webpage generation, systems such as UICopilot~\cite{gui2025uicopilot}, ScreenCoder~\cite{jiang2025screencoder}, and DesignCoder~\cite{chen2025designcoder} adopt hierarchical pipelines that convert screenshots into layouts and then into executable code.
WebGen-Agent~\cite{lu2025webgen} further incorporates visual feedback from rendered pages to iteratively improve generation quality. Although these methods enhance code correctness or layout reconstruction, their hierarchies are still limited to reasoning or code granularity. In contrast, we define hierarchy at the design abstraction level, representing a shift from code-centric orchestration to design-abstraction-driven multimodal generation with structured cross-modal refinement.

\subsection{Webpage Generation Benchmark.}
While there are several benchmarks in the web UI domain, few evaluate text-to-web generation with native multimodal asset creation. Existing datasets and evaluation suites typically fall into three categories. First, strictly code-centric benchmarks focus on HTML/CSS correctness without considering visual content~\cite{yun2024web2code}. Second, image-to-code benchmarks evaluate the reconstruction of webpages from screenshots, emphasizing layout fidelity rather than intent-driven multimodal generation~\cite{lu2025webgen,chen2025designcoder,gui2025webcode2m,awal2025webmmu}. Third, some tasks provide static image assets to be placed as placeholders, largely ignoring the quality and consistency of generated content~\cite{wang2025webgen}. Consequently, none of the existing benchmarks adequately assess the alignment between generated native assets and global page semantics. This gap motivates the introduction of \ourbenchmark{}, providing a systematic framework to evaluate fine-grained multimodal webpage quality.

\section{Method}

\begin{figure*}[!htbp]
    \centering
    \includegraphics[width=\linewidth]{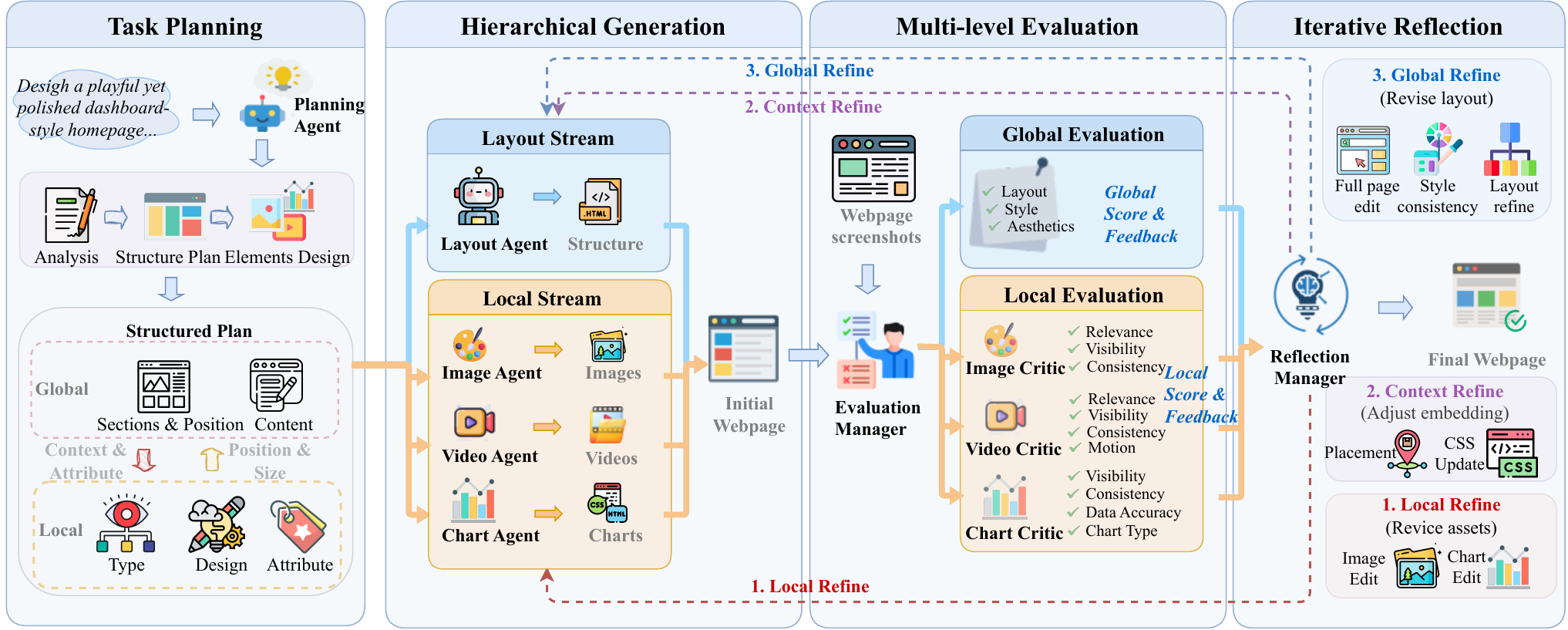}
    \caption{\textbf{An overview of the proposed framework \ourmethod{}.} The framework generates webpages through four key steps: Task planning, hierarchical generation, multi-level evaluation and iterative reflection.}
    \label{fig:framework}
\end{figure*}

Inspired by the workflow of human designers, \ourmethod{} models webpage generation as a hierarchical, structured process that performs hierarchical planning, element-wise generation (Sec.~\ref{sec:method_planning}), followed by hierarchical self-reflection to iteratively refine content (Sec.~\ref{sec:method_reflection}), and is evaluated using multi-level criteria (Sec.~\ref{sec:method_benchmark}).
An overview of the framework is shown in Fig.~\ref{fig:framework}.

\subsection{Hierarchical Planning and Generation}
\label{sec:method_planning}

The planning stage organizes webpage generation into two levels: a global layout plan and local element plans. \ourmethod{} first constructs a global layout plan that defines the section hierarchy, spatial organization, and page-level style attributes, and then derives local element plans conditioned on this global context, specifying each component’s role, layout constraints, and style guidance.

\noindent
\textbf{Global Layout Planning.}
The global layout plan defines the overall structure of the webpage, including section hierarchy, ordering, and spatial organization. For each section, it specifies both the layout of elements (e.g., positions and approximate sizes) and their intended content, such as titles, paragraphs, images, or charts.
Beyond abstract structure, the global plan also introduces explicit placeholders for multimodal elements, annotating their intended positions,  sizes, and layout constraints.
By embedding such local element priors into the global layout, the planner ensures that multimodal components are natively integrated into the page structure.

\noindent
\textbf{Local Element Planning.}
For each multimodal element specified in the global layout, the planner constructs a corresponding local plan to guide its content generation.
Each local plan is grounded in the global context and includes two types of information: \textit{\textbf{i) context information}}, such as the webpage section, the functional role of the element, and the overall page style; and \textit{\textbf{ii) meta attributes}}, which describe modality-specific properties such as visual style, color tone, motion, or specific data requirements.
The local plan also specifies which generation tool should be invoked for the element.
During generation, both the context information and meta attributes are provided as inputs to the corresponding generator.
This design allows local generators to operate in parallel while remaining aligned with the global design intent, ensuring stylistic and functional consistency across modalities.

\noindent
\textbf{Plan Execution.}
Once the plans are constructed, each component is executed by its corresponding generator.
The global layout plan is first converted into the HTML/CSS structure of the webpage, creating sections and placeholders for multimodal elements.
Each local element plan is then executed by the designated generation tool to produce the corresponding asset (e.g., image, video, or chart) according to its context and meta attributes.
The generated assets are subsequently inserted into the webpage to assemble the complete webpage.

\subsection{Hierarchical Self Reflection}
\label{sec:method_reflection}
After designing an initial draft of a webpage, human designers typically refine the design through iterative adjustments.
Inspired by this process, \ourmethod{} implements a hierarchical self-reflection mechanism that iteratively improves the generated webpage at three complementary levels: \textit{local}, \textit{context}, and \textit{global}.

\noindent
\textbf{Local refine.} Human designers typically begin refinement by inspecting individual assets, ensuring that each visual element is semantically correct and visually sound. Following this principle, \ourmethod{} first improves the intrinsic quality of each multimodal element, such as images or charts. The system evaluates each element to identify potential visual or semantic issues and generates corresponding refinement instructions: for images, this may involve inpainting, color adjustment, or object correction, while for charts, it may include fixing labels, axes, or legends. These instructions are then executed via specialized agents, such as image editing models or localized HTML/CSS updates, ensuring that each component meets quality and consistency standards before integration.

\noindent
\textbf{Context refine.}
Even when individual elements are visually and semantically correct, their integration into the surrounding layout can introduce issues such as misalignment, clipping, or inconsistent spacing. Context Refine addresses these problems by analyzing the relevant HTML snippets and generating context-aware adjustments. These are applied through targeted structural edits, such as CSS patches, block resizing, or snippet replacement, ensuring that each element aligns harmoniously with its surroundings and maintains both visual consistency and spatial coherence across the page.

\noindent
\textbf{Global refine.} After local and context-level refinements, the system evaluates the entire webpage to detect high-level layout and style inconsistencies, using both the HTML code and the rendered screenshot as references.
Global refine performs targeted edits to the HTML and page structure, enforcing consistent layout, spacing, and visual style across all sections.
This holistic refinement ensures improved visual balance, structural coherence, and overall alignment with the intended design.

\begin{figure*}[!tbp]
    \centering
    \includegraphics[width=\linewidth]{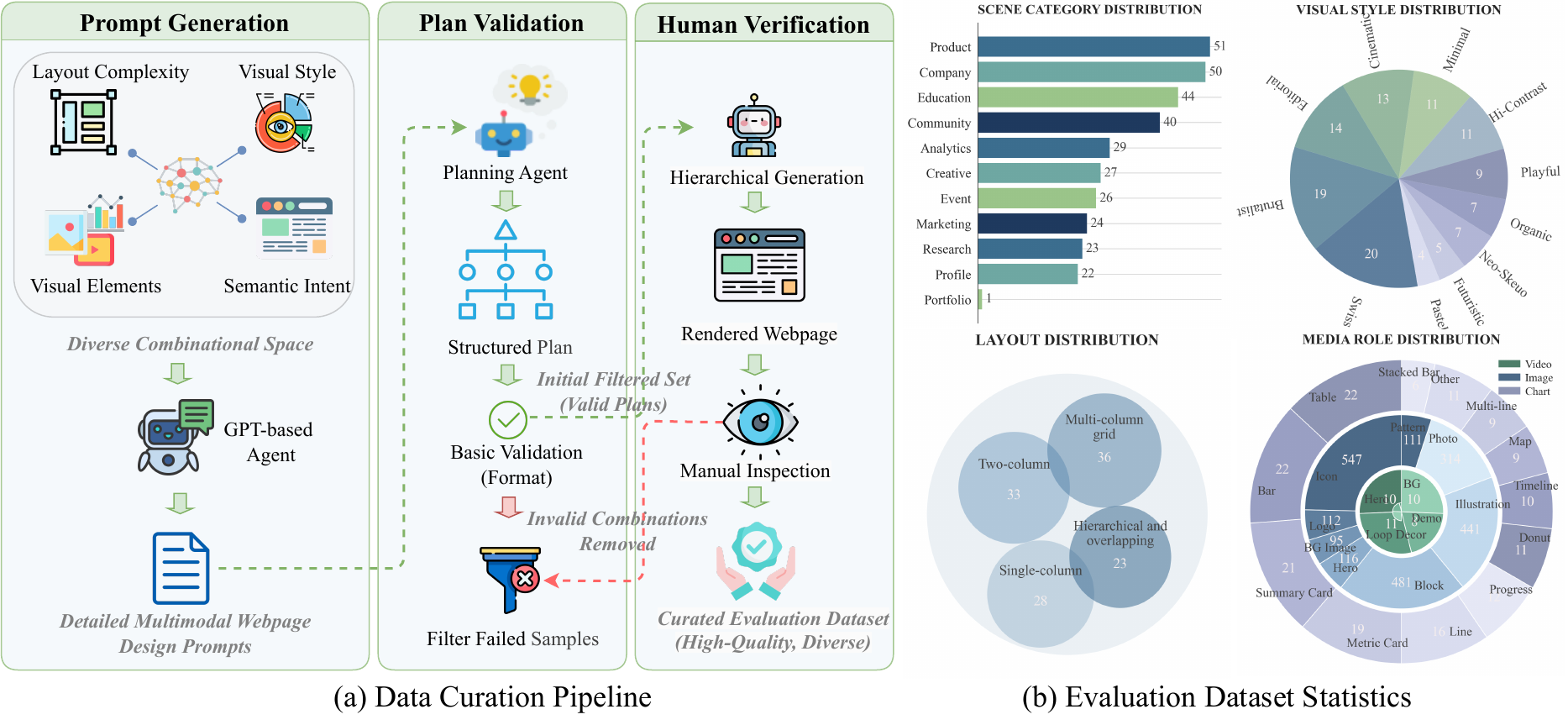}
    \caption{\textbf{Overview of \ourbenchmark{}. (a) Dataset construction process}, including data generation controlled by layout complexity, visual style, semantic intent, and multimodal elements, followed by a filtering pipeline with automatic format validation and manual quality control. \textbf{(b) Statistical summary of the final evaluation set}, consisting of \textbf{120} webpages spanning \textbf{11} scene categories and \textbf{11} visual styles, and featuring diverse multimodal compositions, including \textbf{4} types of videos, \textbf{8} types of images, and \textbf{17} types of charts.}
    \label{fig:dataset_pipeline}
\end{figure*}

\subsection{\ourbenchmark{}}
\label{sec:method_benchmark}
\subsubsection{Evaluation Dataset.}
To evaluate multimodal webpage generation models, we construct \ourbenchmark{}, a curated benchmark reflecting realistic and diverse webpage designs. As illustrated in Fig.~\ref{fig:dataset_pipeline}(a), we first generate a large pool of webpage design prompts through a two-step process. We begin by randomly sampling values along four key dimensions: \textit{layout complexity} (e.g., single-column, multi-column grid, hierarchical layouts), \textit{visual style} (e.g., minimal, editorial, playful), \textit{multimodal elements} (e.g., text, images, videos, charts), and \textit{semantic intent} (e.g., landing pages, dashboards, portfolios). These sampled values define a structured scenario requirement. In the second step, a MLLM agent expands this scenario into a detailed prompt describing a complete webpage design, including its content, structure, and stylw.

Each generated prompt is then converted into a structured generation plan and subjected to automatic format validation, serving as an initial quality control stage. The corresponding webpages are subsequently rendered and manually inspected. Samples exhibiting implausible layouts, inconsistent visual styles, or unrealistic combinations of multimodal elements are discarded, ensuring that the final benchmark contains high-quality, diverse webpages suitable for evaluation.

The remaining high-quality samples constitute the final evaluation set, comprising \textbf{120} carefully curated webpages. Fig.~\ref{fig:dataset_pipeline}(b) summarizes \ourbenchmark{} statistics, highlighting its diversity. First, the dataset covers webpages with varied intents, including informational, analytical, creative, and commercial use cases (Fig.~\ref{fig:dataset_pipeline}(b, top left)). Second, these webpages exhibit a wide range of visual styles, from clean, text-oriented designs (e.g., Swiss-style) to expressive, visually rich aesthetics (e.g., brutalist, cinematic) (Fig.~\ref{fig:dataset_pipeline}(b, top right)). Third, in terms of structure, pages vary in complexity, ranging from simple single-column layouts to multi-column and hierarchical compositions (Fig.~\ref{fig:dataset_pipeline}(b, bottom left)). Fourth, \ourbenchmark{} incorporates diverse multimodal content, including images, videos, and data visualizations, which fulfill both functional and aesthetic roles within a page (Fig.~\ref{fig:dataset_pipeline}(b, bottom right)). This diversity ensures broad coverage of real-world webpage designs and provides a balanced evaluation set across scenarios, structures, and multimodal compositions.

\subsubsection{Multi-level Evaluation.}
Evaluating webpages is inherently challenging due to the interplay between global layout, local content, and diverse embedded elements. To comprehensively evaluate these dimensions, we introduce a multi-level evaluation scheme that decomposes quality assessment into global- and local-level criteria,  enabling structured measurement of both overall page quality and embedded multimodal components.

\textit{Global level evaluation} defines a set of metrics for assessing the overall quality of a webpage, focusing on three key dimensions: 1) \textit{layout correctness} evaluates whether the section hierarchy, ordering, and spatial arrangement of elements conform to the structure implied by the user’s design prompt; 2) \textit{style coherence} measures the consistency of visual attributes such as color palette or overall design theme across all sections of the page; and 3) \textit{aesthetic quality} captures the visual balance, readability, and harmony of the rendered webpage, reflecting its overall appeal and user experience. By combining these dimensions, global evaluation provides a structured assessment of both functional layout and holistic visual presentation.

\textit{Local level evaluation} assesses the quality of individual multimodal elements embedded within the webpage, including images, videos, and charts.
Each element is examined both for its intrinsic quality and for how well it integrates with the surrounding layout and the overall page style.
For images and videos, the evaluation considers semantic relevance, visual or motion characteristics, and how naturally the asset fits its intended role within the page.
For charts, it assesses the clarity and accuracy of data presentation, as well as the consistency with the overall page design.
The evaluation also explicitly accounts for missing or incomplete elements, treating the absence of components implied by the user prompt as critical failures at the local level.
These criteria provide a detailed, element-level assessment that complements global evaluation and enables systematic analysis of multimodal webpage content.

To convert the qualitative evaluation into quantitative scores, we design two complementary scoring strategies for different evaluation dimensions.
For dimensions that involve multiple compositional criteria, such as layout correctness and style coherence, we employ a \textit{penalty-based scoring mechanism}.
The evaluator identifies all violations according to predefined rules and assigns a penalty to each issue based on its severity.
The final score for a sample is computed as
\begin{equation}
score = \max\left(0, 1 - \alpha \cdot \sum_i p_i \right),
\end{equation}
where $p_i$ denotes the penalty associated with the $i$-th detected issue and $\alpha$ is a normalization factor that controls the overall penalty strength.

For dimensions that require more holistic judgment, such as \textit{aesthetic quality} and the quality of local multimodal elements (e.g., images, videos, and charts), we adopt a \textit{graded scoring scheme}.
Each item is assigned a score from a discrete scale $\{0, 0.2, 0.4, 0.6, 0.8, 1.0\}$ to reflect different quality levels.
For each dimension, the final benchmark score is obtained by averaging the scores of all samples, producing values in the range $[0,1]$.
The overall performance of a model is then summarized by averaging the scores across all evaluation dimensions.

\section{Experiments}
\subsection{Experimental Setup}
The hierarchical planner is implemented using GPT-5.1~\cite{openai_gpt51}, which produces structured plans for webpage layout and multimodal elements.
For content generation, images are generated using GPT-Image-1~\cite{openai_gpt_image_1}, videos are generated using the OpenAI video model (Sora-2)~\cite{openai_sora_2}, and charts are generated as executable ECharts-based HTML by OpenAI-GPT-5.1.
Hierarchical reflection is enabled by default, with OpenAI-GPT-5.1 serving as the judger.
During reflection, global layout and chart components are revised using OpenAI-GPT-5.1, while image components are refined using GPT-Image-1 (edit).
Reflection proceeds until convergence or a maximum of 3 iterations.

We compare \ourmethod{} with both code generation-based and agent-based baselines on \ourbenchmark{}.
Code generation-based methods generate webpages in an end-to-end code generation paradigm, while agent-based baselines are implemented using  \texttt{bolt.diy}~\cite{stackblitz_bolt_diy} or \texttt{Openhands}~\cite{wang2024openhands}.
The evaluated models include: OpenAI-GPT 4o~\cite{openai_gpt4o}, OpenAI-GPT 5mini~\cite{openai_gpt5mini}, OpenAI-GPT 5~\cite{openai_gpt5},
 OpenAI-GPT~5.1~\cite{openai_gpt51},
Qwen2.5-Coder-7B-Inst~\cite{hui2024qwen25coder},
Qwen2.5-Coder-32B-Inst~\cite{hui2024qwen25coder},
Qwen3-Coder-30B-A3B-Inst~\cite{qwen3technicalreport},
and Qwen2.5-72B-Inst~\cite{qwen2.5}, and Gemini-2.5-Pro~\cite{comanici2025gemini}. All evaluations are conducted three times and reported as the mean and standard deviation.

\subsection{Main Results}

\noindent
\textbf{Paradigm Comparison on \ourbenchmark{}.}
We evaluate \ourmethod{} under different webpage generation paradigms on \ourbenchmark{}, including
\textit{code-only one-shot generation},
\textit{code-only agent-based generation},
and \textit{multimodal web agent generation}.
Table~\ref{tab:main} reports performance across six evaluation dimensions, where layout, style, and aesthetics measure global page-level quality, while image, video, and chart assess the quality and integration of local multimodal elements.
\ourmethod{}, which enables agentic coordination with native multimodal asset generation, achieves the best performance on both global and local metrics, with an average score of 0.75.
In particular, it shows substantial improvements on multimodal element metrics, including image, video, and chart.
These results highlight the limitation of code-only generation pipelines and demonstrate the advantage of treating multimodal content generation as a first-class action within the agent loop.

\begin{table*}[!t]
\centering
\caption{\textbf{Comparison on \ourbenchmark{}.}
We compare three paradigms: \textbf{(i) Code-only One-shot} (end-to-end HTML/CSS generation),
\textbf{(ii) Code-only Agents} (agentic execution but restricted to code-only assets),
and \textbf{(iii) Multimodal Web Agents} that can invoke AIGC tools to generate/edit multimodal assets.
\textbf{Code-only Agent} baselines are implemented with \texttt{bolt.diy}~\cite{stackblitz_bolt_diy} and \texttt{Openhands}~\cite{wang2024openhands}, where multimodal contents are typically represented by code-based placeholders (e.g., links or SVG). \textit{\ourmethod{} instead invokes multimodal AIGC tools to generate and refine assets, achieving significantly better results.}
\textbf{Bold} and \underline{underline} indicate the best and second-best performance, respectively. Our method is highlighted blue for clarity.}
\label{tab:main}
\resizebox{0.98\linewidth}{!}{%
\begin{tabular}{lccccccc}
\toprule
\multirow{2}{*}{\textbf{Method}}
& \multicolumn{3}{c}{\textbf{Global}}
& \multicolumn{3}{c}{\textbf{Local}}
& \multirow{2}{*}{\textbf{Average}} \\
\cmidrule(lr){2-4} \cmidrule(lr){5-7}
& Layout & Style & Aesthetics & Image & Video & Chart & \\
\midrule

\multicolumn{8}{l}{\textit{\textbf{(I) Code-only One-shot}}} \\ \midrule
Qwen2.5-Coder-7B-Instruct      & 0.01\std{0.00} & 0.00\std{0.00} & 0.78\std{0.00} & 0.41\std{0.01} & 0.00\std{0.00} & 0.24\std{0.01} & 0.24\std{0.00} \\
Qwen2.5-Coder-32B-Instruct     & 0.09\std{0.01} & 0.03\std{0.00} & 0.84\std{0.00} & 0.39\std{0.00} & 0.02\std{0.00} & 0.28\std{0.03} & 0.27\std{0.01} \\
Qwen3-Coder-30B-A3B-Instruct   & 0.13\std{0.01} & 0.15\std{0.00} & 0.57\std{0.00} & 0.08\std{0.00} & 0.00\std{0.00} & 0.25\std{0.02} & 0.20\std{0.00} \\
Qwen2.5-72B-Instruct           & 0.10\std{0.00} & 0.02\std{0.01} & 0.82\std{0.00} & 0.40\std{0.01} & 0.00\std{0.00} & 0.25\std{0.02} & 0.27\std{0.00} \\
Gemini-2.5-Pro                 & 0.57\std{0.00} & 0.24\std{0.02} & 0.94\std{0.00} & 0.43\std{0.01} & 0.00\std{0.00} & 0.45\std{0.01} & 0.44\std{0.00} \\
OpenAI-GPT-4o                  & 0.02\std{0.00} & 0.05\std{0.02} & 0.48\std{0.00} & 0.06\std{0.00} & 0.00\std{0.00} & 0.02\std{0.00} & 0.11\std{0.00} \\
OpenAI-GPT-5mini               & 0.63\std{0.04} & 0.40\std{0.02} & 0.95\std{0.00} & 0.21\std{0.01} & 0.00\std{0.00} & 0.50\std{0.01} & 0.45\std{0.01} \\
OpenAI-GPT-5                   & 0.78\std{0.04} & 0.40\std{0.03} & \underline{0.96}\std{0.00} & 0.14\std{0.00} & 0.02\std{0.00} & \underline{0.52}\std{0.04} & 0.47\std{0.02} \\
OpenAI-GPT-5.1                 & 0.73\std{0.01} & 0.44\std{0.00} & \underline{0.96}\std{0.00} & 0.05\std{0.00} & 0.00\std{0.00} & 0.35\std{0.01} & 0.42\std{0.00} \\
\midrule

\multicolumn{8}{l}{\textit{\textbf{(II) Code-only Agents}}} \\ \midrule
\rowcolor[HTML]{EFEFEF} \multicolumn{8}{l}{i) Bolt.diy} \\
Qwen2.5-Coder-7B-Instruct      & 0.02\std{0.00} & 0.03\std{0.00} & 0.77\std{0.00} & 0.36\std{0.00} & 0.00\std{0.00} & 0.23\std{0.01} & 0.23\std{0.00} \\
Qwen2.5-Coder-32B-Instruct     & 0.08\std{0.00} & 0.02\std{0.00} & 0.85\std{0.00} & 0.48\std{0.00} & 0.02\std{0.01} & 0.31\std{0.00} & 0.29\std{0.00} \\
Qwen3-Coder-30B-A3B-Instruct   & 0.12\std{0.03} & 0.07\std{0.01} & 0.71\std{0.00} & 0.15\std{0.00} & 0.00\std{0.00} & 0.32\std{0.02} & 0.23\std{0.01} \\
Qwen2.5-72B-Instruct           & 0.07\std{0.01} & 0.03\std{0.00} & 0.83\std{0.00} & 0.31\std{0.01} & 0.05\std{0.02} & 0.30\std{0.01} & 0.26\std{0.00} \\
Gemini-2.5-Pro                 & 0.63\std{0.01} & 0.24\std{0.02} & 0.93\std{0.00} & 0.38\std{0.01} & 0.00\std{0.00} & 0.50\std{0.01} & 0.45\std{0.00} \\
OpenAI-GPT-4o                  & 0.04\std{0.00} & 0.02\std{0.00} & 0.85\std{0.00} & 0.21\std{0.01} & 0.00\std{0.00} & 0.12\std{0.01} & 0.21\std{0.00} \\
OpenAI-GPT-5mini               & 0.67\std{0.01} & 0.36\std{0.03} & 0.95\std{0.00} & 0.12\std{0.01} & 0.00\std{0.00} & 0.48\std{0.01} & 0.43\std{0.01} \\
OpenAI-GPT-5                   & 0.77\std{0.02} & 0.43\std{0.05} & 0.95\std{0.00} & 0.06\std{0.00} & 0.00\std{0.00} & 0.50\std{0.01} & 0.45\std{0.01} \\
OpenAI-GPT-5.1                 & 0.74\std{0.04} & 0.39\std{0.01} & \underline{0.96}\std{0.00} & 0.30\std{0.00} & 0.00\std{0.00} & 0.36\std{0.02} & 0.46\std{0.01} \\ \hline

\rowcolor[HTML]{EFEFEF} \multicolumn{8}{l}{ii) OpenHands} \\
Gemini-2.5-Pro                 & 0.43\std{0.03} & 0.21\std{0.00} & 0.93\std{0.00} & 0.31\std{0.01} & 0.00\std{0.00} & 0.47\std{0.00} & 0.39\std{0.00} \\
OpenAI-GPT-4o                  & 0.03\std{0.01} & 0.02\std{0.00} & 0.83\std{0.00} & 0.11\std{0.00} & 0.00\std{0.00} & 0.04\std{0.01} & 0.17\std{0.00} \\
OpenAI-GPT-5mini               & 0.60\std{0.00} & 0.31\std{0.01} & 0.94\std{0.00} & 0.05\std{0.01} & 0.00\std{0.00} & 0.47\std{0.01} & 0.39\std{0.00} \\
OpenAI-GPT-5                   & 0.76\std{0.03} & 0.41\std{0.04} & 0.95\std{0.00} & 0.02\std{0.00} & 0.00\std{0.00} & 0.49\std{0.01} & 0.44\std{0.00} \\
OpenAI-GPT-5.1                 & 0.74\std{0.00} & 0.39\std{0.03} & \underline{0.96}\std{0.00} & 0.30\std{0.00} & 0.00\std{0.00} & 0.36\std{0.01} & 0.46\std{0.01} \\
\midrule

\multicolumn{8}{l}{\textit{\textbf{(III) Multimodal Web Agents (with AIGC tools)}}} \\ \midrule
\rowcolor{blue!8}
\multicolumn{8}{l}{\textbf{\ourmethod{} (Ours)}} \\
Gemini-2.5-Pro                 & 0.68\std{0.01} & 0.35\std{0.01} & \underline{0.96}\std{0.00} & 0.81\std{0.00} & 0.57\std{0.02} & 0.43\std{0.00} & 0.63\std{0.00} \\
OpenAI-GPT-4o                  & 0.16\std{0.00} & 0.10\std{0.01} & 0.86\std{0.00} & 0.42\std{0.02} & 0.29\std{0.00} & 0.32\std{0.03} & 0.36\std{0.01} \\
OpenAI-GPT-5mini               & 0.73\std{0.01} & 0.42\std{0.01} & 0.95\std{0.00} & 0.84\std{0.00} & \underline{0.63}\std{0.05} & 0.50\std{0.01} & 0.68\std{0.01} \\
OpenAI-GPT-5                   & \textbf{0.85}\std{0.02} & \underline{0.53}\std{0.04} & \textbf{0.97}\std{0.00} & \underline{0.86}\std{0.01} & 0.52\std{0.03} & \textbf{0.54}\std{0.04} & \underline{0.71}\std{0.00} \\
OpenAI-GPT-5.1                 & \underline{0.83}\std{0.02} & \textbf{0.54}\std{0.03} & \textbf{0.97}\std{0.00} & \textbf{0.88}\std{0.00} & \textbf{0.75}\std{0.02} & \textbf{0.54}\std{0.02} & \textbf{0.75}\std{0.01} \\
\bottomrule
\end{tabular}
}
\end{table*}

\noindent
\textbf{Comparison on WebGen-Bench~\cite{lu2025webgenbench}.}
To provide a broader perspective, we evaluated WebGen-Bench~\cite{lu2025webgenbench}, which primarily tests functional backend code, logic, and component completeness. Because the user prompts in this task lack specific visual instructions, the ``Appearance Score'' does not reflect the content-generation capabilities we focus on. Additionally, our agent is not explicitly designed for backend code generation. Despite these disadvantages, \ourmethod{} still achieved highly competitive results, as shown in Table~\ref{tab:com_webgen_bench}.

\begin{table}[!htbp]
\centering
\caption{\textbf{Comparison on WebGen-Bench~\cite{lu2025webgenbench}.} The best Accuracy and Appearance Score are highlighted in \textbf{Bold}.}
\label{tab:com_webgen_bench}
\resizebox{0.98\linewidth}{!}{%
\begin{tabular}{lcccccc}
\toprule
\textbf{Method} & \textbf{Yes} & \textbf{Partial} & \textbf{No} & \textbf{Start Failed} & \textbf{Accuracy} & \textbf{Appearance Score} \\ \hline
OpenAI-GPT 5.1 (Code-only) & 42.9 & 9.9 & 47.8 & 0.2 & 47.1 & 1.1 \\
WebGen-LM-32B (Bolt.diy) & 34.2 & 8.0 & 57.8 & 0.0 & 38.2 & 2.8 \\
WebGenAgent-LM-7B & 40.2 & 10.5 & 49.3 & 0.0 & 45.4 & 3.7 \\
OpenAI-GPT 5.1 (Bolt.diy) & 48.7 & 13.4 & 36.2 & 1.7 & \textbf{55.4}& 3.8 \\
OpenAI-GPT 5.1 (OpenHands) & 37.7 & 12.1 & 50.2 & 0.0 & 43.7 & 2.9 \\ \rowcolor{blue!8}
\ourmethod{} + GPT5.1 (Ours) & 47.8 & 15.1 & 36.5 & 0.6 & \textbf{55.4}& \textbf{3.9} \\ \bottomrule
\end{tabular}
}
\end{table}

\begin{table*}[!t]
\centering
\caption{\textbf{Ablation on hierarchical planning and hierarchical reflection.}
\textbf{Planning:} evaluated under \emph{no reflection}.
\textbf{Reflection:} evaluated under \emph{full hierarchical planning} (global + context).
Our method is highlighted blue for clarity.}
\label{tab:ablation_clean}
\resizebox{0.8\linewidth}{!}{%
\begin{tabular}{lccccccc}
\toprule
\multirow{2}{*}{\centering\textbf{Setting}}
& \multicolumn{3}{c}{\textbf{Global}}
& \multicolumn{3}{c}{\textbf{Local}}
& \multirow{2}{*}{\centering\textbf{Avg}} \\
\cmidrule(lr){2-4} \cmidrule(lr){5-7}
& Layout & Style & Aes & Img & Vid & Chart & \\
\midrule
\multicolumn{8}{l}{\textbf{(A) Hierarchical Planning (no reflection)}} \\
\midrule
One-shot (no planning, no reflection)
& 0.73 & 0.44 & 0.96 & 0.05 & 0.00 & 0.35 & 0.42 \\
Hierarchical planning
& 0.68 & 0.37 & 0.96 & 0.85 & 0.65 & 0.44 & 0.66 \\

\midrule
\multicolumn{8}{l}{\textbf{(B) Hierarchical Reflection (with full hierarchical planning)}} \\
\midrule
No reflection
& 0.68 & 0.37 & 0.96 & 0.85 & 0.65 & 0.44 & 0.66 \\
Local reflection only
& 0.71 & 0.37 & 0.95 & 0.87 & 0.68 & 0.48 & 0.68 \\
Local + Context reflection
& 0.73 & 0.43 & 0.96 & 0.87 & 0.69 & 0.54 & 0.70 \\
Global reflection only
& 0.85 & 0.53 & 0.96 & 0.86 & 0.68 & 0.49 & 0.73 \\

\rowcolor{blue!8}
\textbf{All reflections (Default)}
& \textbf{0.83} & \textbf{0.54} & \textbf{0.97}
& \textbf{0.88} & \textbf{0.75} & \textbf{0.54}
& \textbf{0.75} \\
\bottomrule
\end{tabular}
}
\end{table*}

\begin{table}[!htbp]
    \centering
    \caption{\textbf{Ablation on the effect of AIGC tool access.} Results show that AIGC tools alone provide limited benefits, whereas our hierarchical agent framework unlocks their full potential and significantly improves overall performance.}
    \label{tab:ab_aigc_tools}
    \resizebox{0.95\linewidth}{!}{%
    \begin{tabular}{lccccccc}
    \toprule
         \textbf{Method}&  \textbf{Layout} &\textbf{Style} &\textbf{Aesthetics} &\textbf{Image} &\textbf{Video} &\textbf{Chart} &\textbf{Overall}\\ \hline
         Code-only GPT-5.1&
     0.73& 0.44& 0.96& 0.05& 0.00& 0.35&0.42\\
 Code-only GPT-5.1 + AIGC-tools& 0.83& 0.45& 0.94& 0.05& 0.09& 0.36&0.45\\ \rowcolor{blue!8}
 \ourmethod{}(Ours) & \textbf{0.83}& \textbf{0.54}& \textbf{0.97}& \textbf{0.88}& \textbf{0.75}& \textbf{0.54}&\textbf{0.75}\\ \bottomrule
 \end{tabular}
 }
\end{table}

\subsection{Ablation Studies}

\noindent
\textbf{Ablation on Hierarchical planning.}
Table~\ref{tab:ablation_clean}(A) shows that without hierarchical planning, the agent collapses to one-shot generation and fails on multimodal elements, especially images and videos.
Introducing hierarchical planning enables structured coordination of multimodal content and substantially improves performance.
We further ablate local planning by disabling it from the full system, which results in a clear drop in overall performance (Avg: 0.75 → 0.69), with pronounced degradation on local metrics (e.g., Image and Video), confirming the necessity of context-aware local planning.

\noindent
\textbf{Ablation on Hierarchical reflection.}
Table~\ref{tab:ablation_clean}(B) reveals complementary roles of different reflection levels.
Local reflection mainly improves element-level quality, while global reflection primarily enhances layout and style coherence.
Combining all reflection levels yields the best overall performance.

\noindent
\textbf{Ablation on AIGC Tool Access.}
To analyze whether the performance gains of \ourmethod{} primarily stem from the use of AIGC tools themselves, we conduct an ablation study comparing three settings: (1) a standard code-only generation pipeline, (2) the same pipeline augmented with direct access to the identical AIGC tools used in our system, and (3) our full hierarchical agent framework. As shown in Table~\ref{tab:ab_aigc_tools}, simply bolting AIGC tools onto a standard code-generation pipeline yields marginal improvements (Overall: 0.42 to 0.45). It is only through the explicit context-aware planning and multi-level reflection of MM-WebAgent that the overall score jumps to 0.75. This confirms that the performance gains are genuinely driven by our proposed agentic design, and our hierarchical planning are necessary to unlock their full potential.

\begin{wrapfigure}{r}{0.48\textwidth}
\centering
\includegraphics[width=\linewidth]{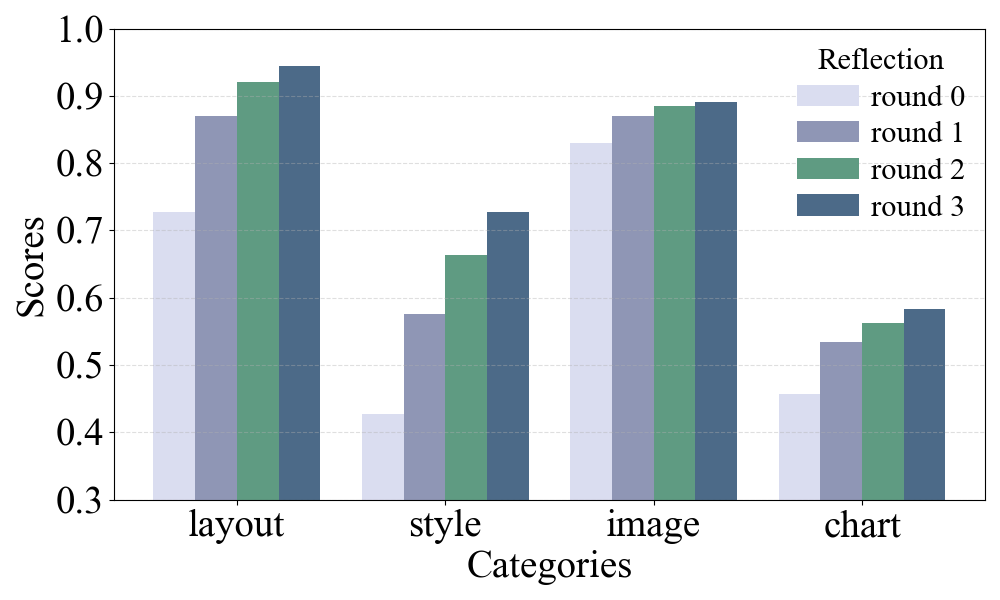}
    \caption{\textbf{Effect of reflection iterations on global and local evaluation metrics.} Hierarchical reflection steadily improves both global and local metrics.}
    \label{fig:vis_ab1_reflection}
\end{wrapfigure}

\noindent
\textbf{Ablation on Reflection iterations.}
Fig.~\ref{fig:vis_ab1_reflection} shows that most gains are achieved within the first few reflection rounds, indicating that hierarchical reflection enables efficient refinement without excessive iterations.

\subsection{Computational Cost}

Table~\ref{tab:compute_costs} reports the average cost and latency per task compared with representative code-centric agents. Although \ourmethod{} involves multiple LLM calls due to its planning, multimodal generation, and reflection stages, the overall runtime remains competitive. In particular, the average execution time of \ourmethod{} (155.8s) is comparable to \texttt{Openhands} (182.4s), despite handling substantially more complex multimodal generation tasks.
While the monetary cost of \ourmethod{} is higher than code-only agents, this increase primarily reflects the intrinsic complexity of native multimodal webpage generation rather than redundant computation. As multimodal models continue to improve and open-source alternatives emerge, the effectiveness of our framework will naturally benefit from these advancements.

\begin{table}[!htbp]
    \centering
    \caption{\textbf{Per-task latency and token comparison with representative code-centric agents}. $\ddagger$: In our implementation, image, video, and chart generation are executed in parallel; thus the overall latency faster than the sum of these modules.}
    \label{tab:compute_costs}
    \resizebox{0.95\linewidth}{!}{%
    \begin{tabular}{ccccccccc}
    \toprule
      \textbf{Method}   & \textbf{Avg Costs (\$)} & \textbf{Avg Time(s)} & \textbf{Planner} & \textbf{Global} & \textbf{Image} & \textbf{Video}& \textbf{Chart} & \textbf{Reflection}\\ \hline
       \texttt{Openhands}~\cite{wang2024openhands}  & 0.27 & 182.4 & -& -& -& -& -& -\\
        \texttt{bolt.diy}~\cite{stackblitz_bolt_diy}  & 0.14 & 76.9 & -& -& -& -& -& -\\ \rowcolor{blue!8}
        \ourmethod{}  & 3.21 & 155.8$^\ddagger$& 56.1& 58.6& 44.1& 61.1 & 30.8& 41.1\\ \bottomrule
    \end{tabular}
    }
\end{table}

\begin{figure*}[t]
    \centering
    \includegraphics[width=\linewidth]{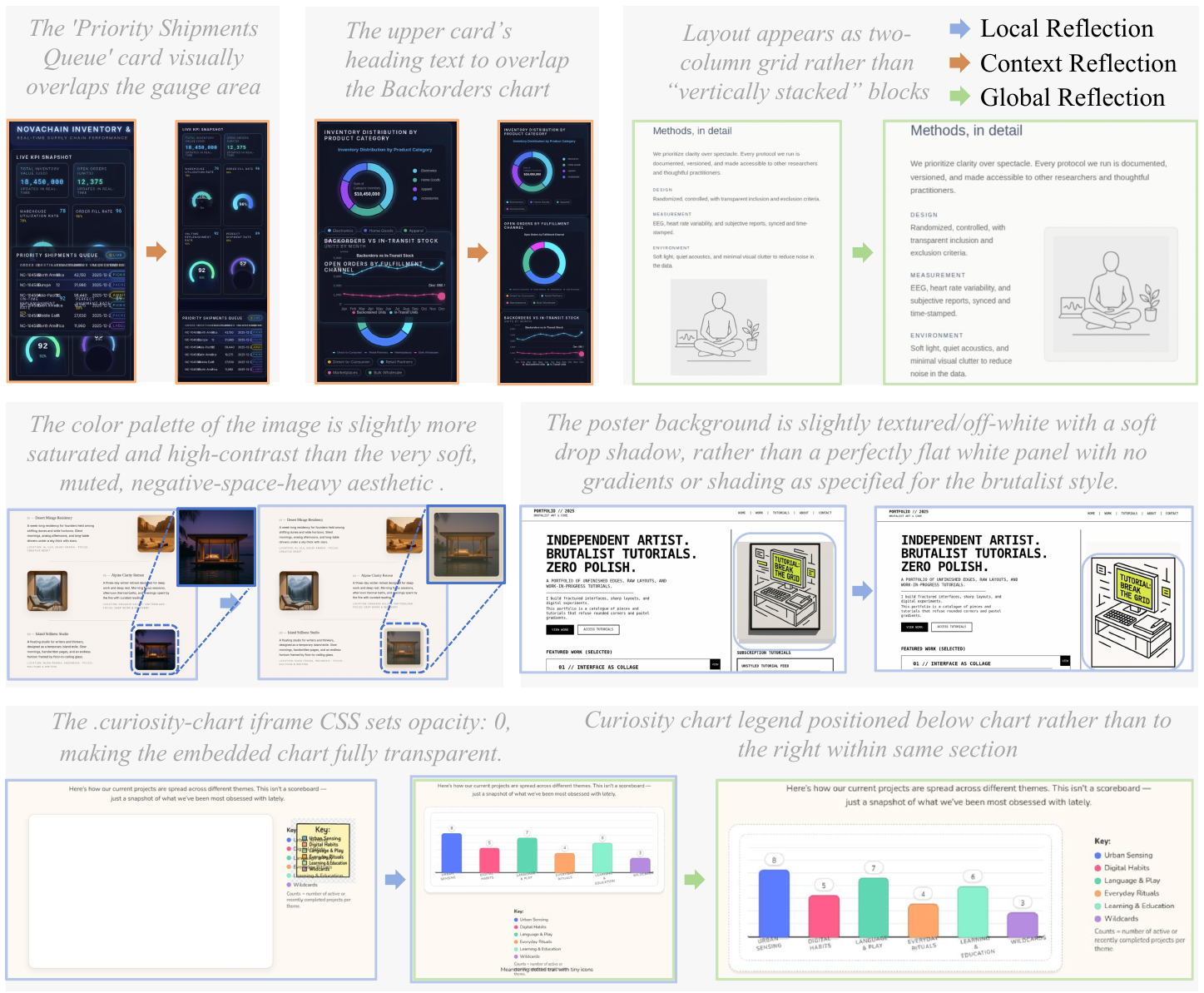}
    \caption{\textbf{Visualization of the hierarchical reflection process.}
    \ourmethod{} progressively refines local multimodal elements and global layout through iterative reflection, with examples of global layout refinement, context refinement (\textit{first row}), local element refinement (\textit{second row}), and local-to-global correction (\textit{third row}).}
    \label{fig:vis_reflection_examples}
\end{figure*}

\subsection{User Study}
To evaluate the agreement between human preferences and our automatic evaluator, we conduct a pairwise user study. This evaluation was conducted with the participation of 50 annotators. All annotators have backgrounds in web design, computer science, or multimodal content creation, enabling them to objectively assess visual quality, layout rationality, and multimodal content integration of the webpages.
During the evaluation, the generated webpages for each task were presented in anonymized and randomly shuffled order. Annotators performed blind assessments without knowing which method produced each result. All evaluators scored the webpages according to predefined criteria (e.g., visual quality, layout coherence, etc.). The final results were obtained by aggregating the scores across all annotators.

For each comparison, participants are shown two webpages generated by different methods, and are asked to compare the webpages in terms of layout quality, content relevance, multimodal asset quality, and the embedding quality of local elements.
Ratings are given on a five-level scale: \emph{much worse}, \emph{worse}, \emph{similar}, \emph{better}, and \emph{much better}. Each response was then mapped to win, tie, or lose depending on whether our method was preferred. The winning rate is computed as the ratio of the number of wins to the total number of pairwise comparisons. Overall, \ourmethod{} achieves a winning rate of \textbf{78.99\%}, indicating that human evaluators strongly prefer webpages generated by our approach compared with the competing methods.

\subsection{Qualitative Results}

Fig.~\ref{fig:vis_comparison} presents qualitative comparisons of webpages generated on \ourbenchmark{} by \ourmethod{} and representative baseline methods.
While baseline approaches often produce incomplete or poorly integrated multimodal elements, \ourmethod{} generates webpages with more coherent layouts, consistent visual styles, and better-aligned multimodal content.
In particular, our method more reliably integrates images and charts into the overall page structure, which better align with the intended design and semantic requirements.

Fig.~\ref{fig:vis_reflection_examples} illustrates the hierarchical reflection behavior of \ourmethod{}. The agent is able to iteratively refine the global layout through global reflection, while local reflection focuses on adjusting individual multimodal elements to better align with the overall webpage style. Moreover, local reflection can propagate to the global level, leading to more coherent overall page structures.

\section{Conclusion}

We present \ourmethod{}, a hierarchical framework for multimodal webpage generation that integrates structured planning, hierarchical generation, and iterative self-reflection. The planning stage organizes the global layout and specifies local elements, enabling the generation of diverse multimodal content, while hierarchical reflection iteratively adjusts both local elements and global layouts to enhance overall consistency and visual quality. To evaluate performance in generating diverse and coherent multimodal webpages, we introduce \ourbenchmark{}, a benchmark encompassing a wide range of layouts, visual styles, and multimodal compositions. Experiments show that \ourmethod{} outperforms both code generation-based and agent-based baselines, demonstrating its effectiveness in generating well-integrated multimodal webpage.

\section{Limitation and Future Work}
Our approach relies on external AIGC tools for generating images, videos, and charts, making webpage quality susceptible to tool-level limitations such as instability, bias, safety filters, or changes in availability. Our framework also assumes a fixed set of tools and invocation patterns, restricting flexibility in dynamic tool selection and composition.
Additionally, \ourmethod{} adopts an orchestration-based, training-free agentic formulation. Although this choice allows us to clearly study the impact of hierarchical planning and reflection, it does not leverage learning-based optimization of agent behaviors. Incorporating reinforcement learning or other learning paradigms to optimize planning, tool usage, and reflection strategies over long-term interactions may further improve performance and generalization.

\bibliographystyle{splncs04}
\bibliography{main}

\clearpage
\appendix
\renewcommand{\theHsection}{appendix.\Alph{section}}
\renewcommand{\theHsubsection}{appendix.\Alph{section}.\arabic{subsection}}
\section*{Supplementary Material}

\section{More Qualitative Results}
\begin{figure*}[!h]
    \centering
    \includegraphics[width=0.9\linewidth]{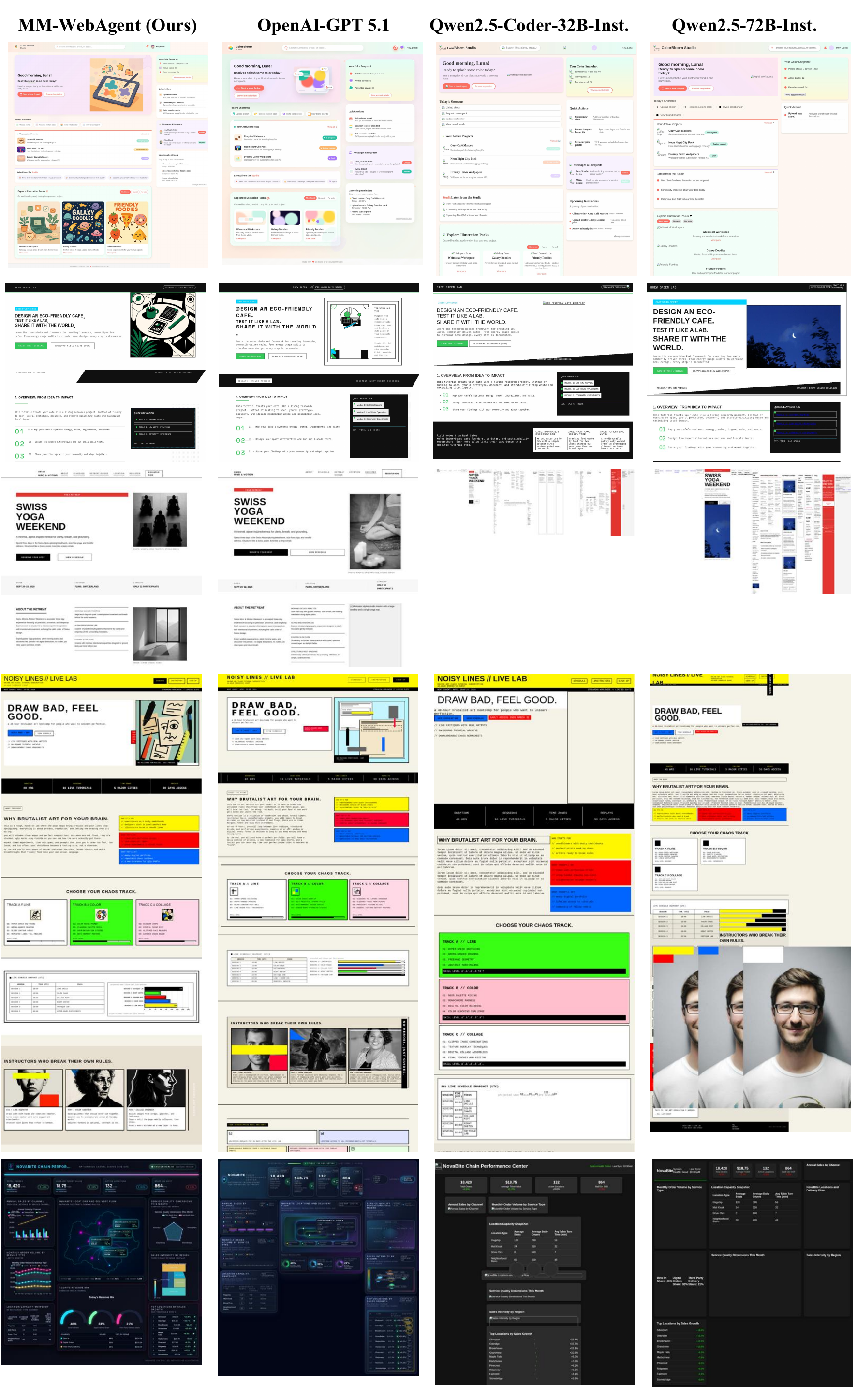}
    \caption{\textbf{More rendered webpage examples generated by \ourmethod{} and baseline methods on \ourbenchmark{}.}}
    \label{fig:vis_comparison_more}
\end{figure*}

We present more examples of generated webpages in Fig.~\ref{fig:vis_comparison_more}.

\section{Prompt Templates}

\subsection{Planner Prompt}

\begin{promptbox}[label={lst:planner_prompt}]
{Planner Prompt for the Webpage-Generation Planning Agent}
You are a webpage-generation planning agent.

The user will give you a prompt describing a webpage they want to generate.
Your task is to analyze the prompt and decompose it into a structured plan for tool calls.

---

### Supported Tools
1. **code_generation** — use to generate the HTML layout of the webpage.
2. **image_generation** — used after layout planning, automatically extracted from all referenced image placeholders.
3. **video_generation** — used to generate videos based on detailed visual descriptions.
4. **data_visualization** — used when the user provides a dataset that needs to be visualized; generates an echart html file and integrates it into the webpage.

---

### Planning Guidelines
1. **HTML Layout & Visual Planning**  
    Write a detailed description for the `code_generation` tool.  
    - Describe all sections of the webpage (e.g., hero banner, logo, navigation bar, background, icons, gallery, footer, etc.).  
    - For any visual element that needs an image (`.png`), video (`.mp4`), or chart (`.html`), insert a clear reference specifying **both the file path and a reasonable layout size (width and height in pixels or % of viewport)** to fit the overall webpage design. For example:
        - `"Hero image shows a cozy café interior at sunrise (path: assets/hero_cafe.png, width: 1200px, height: 600px)"`  
        - `"Sales chart comparing 2022 revenue across cities (path: assets/chart_sales.html, width: 600px, height: 400px)"`  
    - **For multiple charts, generate each chart as a separate HTML file** and reference them individually.  
    - Include layout structure, typography, color palette, and mood.
    - The reference paths must use the format: `.png` for images, `.mp4` for videos, and html for charts.

2. **Image Extraction**  
   After defining the full webpage plan, extract all referenced `(path: assets/xxx.png)` entries and generate corresponding image descriptions for `image_generation`:
    Each entry must include:
    - `"save_path"` consistent with the code reference, all images should be saved `.png` format
    - **context**
        - `section`: webpage section where the image appears (e.g., hero, feature card)
        - `role`: functional role in layout (background, illustration, accent)
        - `global_style`: overall webpage style (e.g., modern minimal, playful, corporate)
    - **compiled_attributes**
        - `visual_style`: photorealistic, illustration, abstract, UI-style, etc.
        - `color_tone`: muted, vibrant, monochrome, pastel, etc.
        - `composition`: wide shot, centered object, negative space, cropped detail
        - `lighting`: soft natural, studio lighting, flat, high contrast
        The `context` and `compiled_attributes` together form an explicit **local planning layer** guiding image generation.
    - `"prompt"` describing the intended visual
    - `"size"` ("1024x1024", "1024x1536", or "1536x1024") for `image_generation` only

3. **Video Extraction**  
   If the webpage description includes any dynamic or animated visual elements (e.g., background videos, hero section animations), extract these and generate corresponding video descriptions for `video_generation`.  
   Each entry should include:
    - `"save_path"` consistent with the code reference, all videos should be saved `.mp4` format
    - **context**
        - `section`: webpage section where the video is embedded
        - `role`: background loop, hero animation, product showcase
        - `global_style`: overall webpage visual style
    - **compiled_attributes**
        - `visual_style`: cinematic, UI-style, abstract, illustrative
        - `motion_intensity`: low or medium
        - `camera_behavior`: static, slow pan, subtle zoom
        - `loopability`: whether the video should loop seamlessly
    - `"prompt"` describing the intended video content in detail
    - `"seconds"` (4, 8, or 12)
    - `"size"` (720x1280, 1280x720, 1024x1792, or 1792x1024)
    
4. **Data Visualization Extraction**
    If the webpage requires charts based on provided datasets, extract all `(path: assets/xxx.html)` references and generate corresponding chart descriptions under `data_visualization`. Ensure each chart’s visual style aligns seamlessly with the overall webpage design. **A greater variety of chart types** is encouraged to enhance data representation and user engagement.

    Each chart entry must include:

    - `"save_path"` consistent with the code reference (e.g., `assets/chart_sales.html`)
    - **context**
        - `section`: webpage section where the chart is placed
        - `role`: analytical, explanatory, comparative, decorative
        - `global_style`: overall webpage design theme
    - **compiled_attributes**
        - `chart_type`: bar chart, line chart, stacked area chart, radar chart, heatmap, etc.
        - `chart_style`: clean, dense, presentation-oriented, dashboard-style
        - `color_palette`: aligned with webpage colors
        - `visual_emphasis`: which data dimensions should stand out
    - `"prompt"` a detailed description specifying:
        - The chart type (e.g., bar chart, line chart, stacked area chart, radar chart, heatmap, etc.)
        - The intended color palette, typography, and visual aesthetics
        - Any additional configuration, such as labels, legends, axes, annotations, transparency, or animations
    - `"source_data"`: markdown format, the complete dataset content required to visualize the chart **must be included** in this field.

---

### Output Format
Return the plan **strictly as JSON**, following this structure:

```json
{
    "code_generation": [
        {
            "prompt": "Detailed webpage layout description, including inline image references (path: assets/xxx.png) and data visualization references (path: assets/xxx.html)."
        }
    ],

    "image_generation": [
        {
            "save_path": "assets/xxx.png",
            "context": {
                "section": "...",
                "role": "...",
                "page_style": "..."
            },
            "compiled_attributes": {
                "visual_style": "...",
                "color_tone": "...",
                "composition": "...",
                "lighting": "..."
            },
            "prompt": "visual description of the image as referenced in the code_generation step",
            "size": "1024x768",
        }
    ],

    "video_generation": [
        {
            "save_path": "assets/xxx.mp4",
            "context": {
                "section": "...",
                "role": "...",
                "page_style": "..."
            },
            "compiled_attributes": {
                "visual_style": "...",
                "motion_intensity": "...",
                "camera_behavior": "...",
                "loopability": "..."
            },
            "prompt": "detailed visual description of the video content as referenced in the code_generation step",
            "seconds": "8",
            "size": "1280x720"
        }
    ],

    "data_visualization": [
        {
            "save_path": "assets/xxx.html",
            "context": {
                "section": "...",
                "role": "...",
                "page_style": "..."
            },
            "compiled_attributes": {
                "chart_style": "...",
                "chart_type": "...",
                "color_palette": "...",
                "visual_emphasis": "..."
            },
            "prompt": "detailed description including dataset content, chart type, colors, style, and configuration",
            "source_data": "relevant dataset content or summary"
        }
    ]
}
\end{promptbox}

\begin{promptbox}[label={lst:planner_image_context}]
{Image Context Template}
The image will be incorporated into a {page_style} webpage, serving as a {role} image in the {section} section. 
\end{promptbox}

\begin{promptbox}[label={lst:planner_image_attr}]
{Image Attribute Template}
The image should have a {visual_style} visual style, {color_tone} color tone, {composition} composition, and {lighting} lighting.
\end{promptbox}

\begin{promptbox}[label={lst:planner_video_context}]
{Video Context Template}
The video will be embedded in a {page_style} webpage, functioning as a {role} video in the {section} section. 
\end{promptbox}

\begin{promptbox}[label={lst:planner_video_attr}]
{Video Attribute Template}
The video should exhibit a {visual_style} style, {motion_intensity} motion intensity, {camera_behavior} camera behavior, and {loopability}.
\end{promptbox}

\begin{promptbox}[label={lst:planner_chart_context}]
{Chart Context Template}
The chart will appear in a {page_style} webpage, acting as a {role} chart in the {section} section. .
\end{promptbox}

\begin{promptbox}[label={lst:planner_chart_attr}]
{Chart Attribute Template}
The chart should be a {chart_type} chart, with a {chart_style} style, {color_palette} color palette, emphasizing {visual_emphasis}.
\end{promptbox}

\begin{promptbox}[label={lst:planner_layout_prompt}]
{Layout Agent Prompt}
Generate a complete HTML file for the following webpage description.

Requirements:
1. **Output only valid HTML code** — no explanations, comments, or markdown formatting.  
   The output should be directly savable as a `.html` file and openable in a browser.
2. **Strictly preserve image references** — if the description includes any image references in the format **(path: xxx)**,  
   you must use **exactly the same file path** for the corresponding `<img>` elements, `<source>` tags, or CSS background-image URLs.  
   Do not modify, rename, or relocate these paths.
3. The HTML should faithfully represent the layout, structure, and style described in the input prompt, including:
   - Semantic sections (hero, header, footer, gallery, etc.)
   - Visual hierarchy and composition
   - Color palette, font choices, and overall theme
4. Include minimal inline CSS or internal `<style>` tags to make the page visually coherent.

Example:
If the description says:
> "The hero section shows a cozy café interior (path: assets/hero_cafe.png)."

Then the generated HTML **must** include:
```html
<img src="assets/hero_cafe.png" alt="cozy café interior">
``` 
\end{promptbox}

\begin{promptbox}[label={lst:planner_imagent_prompt}]
{Image Generation Agent Prompt}
Generate the image based on the following detailed visual description. The image should be high quality and match the style, composition, and mood described.
\end{promptbox}

\begin{promptbox}[label={lst:planner_chart_prompt}]
{Chart Generation Agent Prompt}
Generate a complete, self-contained HTML file for the following visualization description.

Requirements:
1. **Output only valid HTML code** — no explanations, comments, or markdown formatting.  
   The output must be directly savable as a `.html` file and openable in a browser.
2. Use **ECharts** to render the chart.
3. The chart background must be transparent, so it blends seamlessly when embedded into another webpage.
4. **Do not include any layout elements** — no header, footer, sections, captions, or descriptive text. Only the chart container is needed. For reference, the HTML <head> can look like this:

```html
<head>
  <meta charset="UTF-8" />
  <meta name="viewport" content="width=device-width, initial-scale=1.0"/>
  <title>A Short Title</title>
  <script src="https://cdn.jsdelivr.net/npm/echarts@5/dist/echarts.min.js"></script>
  <style>
    html, body {
      height: 100%;
      margin: 0;
      background: transparent;
      font-family: Arial, Helvetica, sans-serif;
    }
    #chart {
      width: 100%;
      height: 100vh;
    }
  </style>
</head>
````

5. The chart must **occupy the full viewport**, filling the `<iframe>` or container entirely. Remove any margins or padding, and ensure the chart container uses the full width and height of the viewport.
6. Ensure the chart is **responsive**, scaling automatically to fit the container while maintaining aspect ratio.
7. **Double-check that the HTML runs without errors and produces the desired visualization, with all data placed correctly.**
\end{promptbox}

\subsection{Evaluation Prompt}

\begin{promptbox}[label={lst:eval_layout_sys_prompt}]
{Layout Evaluation System Prompt}
You are an evaluator that assesses **the layout quality of a generated webpage** based on:

1. **User design prompt**
2. **Generated HTML code**
3. **Rendered webpage screenshot** (the image input)

Your task is to check whether the webpage layout correctly satisfies the required structure, placement, and relationships implied by the design prompt and the HTML.

You must output **all detected layout issues** and assign penalty values according to the rules below.

# **Layout Penalty Rules**

## **1. Element Presence Errors**

### **Critical elements** (0.5 each)

These significantly affect layout structure. Examples:

* Main title / hero section
* Primary image / hero image
* Navigation bar / sidebar / footer
* Key sections explicitly described in `design_prompt` (e.g., “3-column features”, “form section”)
* Any `<section>` or `<div>` with semantic meaning in HTML

Penalties:

* Missing critical element → **0.5**
* Extra critical element not justified by the prompt → **0.5**

### **Minor elements** (0.3 each)

Examples:

* Buttons, icons, small text blocks
* Badges, tags, small images
* Input fields or labels (unless primary)

Penalties:

* Missing minor element → **0.3**
* Extra minor element → **0.3**

## **2. Positioning & Structural Errors**

These check whether elements appear at the correct **spatial positions** relative to the design prompt **AND** the HTML structure.

Penalties:

* Misplaced minor element → **0.1**
  (e.g., button expected under a card but placed above)
* Misplaced critical element → **0.2**
  (e.g., hero text expected left but appears centered)
* Structural mismatch between prompt and screenshot → **1.0**
  Examples:

  * Expected **2×3 grid** rendered as **1-column list**
  * Expected **sidebar-left** placed on right
  * Expected **top navigation** rendered as bottom navigation
  * Expected **section order** wrong (e.g., features above hero)

> If the HTML structure and screenshot disagree (e.g., HTML defines a grid but screenshot shows stacked layout), still penalize.

## **3. Visual Detail Errors**

Not about aesthetics — strictly layout details.

Penalties:

* Wrong shape of an element (e.g., rectangular vs. rounded) → **0.1**
* Wrong size dominance (e.g., primary image too small, title too small) → **0.1**
* Improper spacing/alignment (clearly off from expected) → **0.1**

# **Output Format (Strict Required Format)**

Your output **MUST** follow exactly this structure:

```
Layout Penalties:
- <Issue>: Penalty--<value>
- <Issue>: Penalty--<value>
...
Total Penalty: <sum>
```

Rules:

* Do NOT include any explanations or extra text.
* Use short, concise issue descriptions.
* List **all** issues found.
* Sum must equal the sum of all listed penalties.


# **Evaluation Method**

When evaluating:

* Compare **design_prompt → HTML → screenshot** consistently.
* Trust screenshot as final truth if there's conflict.
* Penalize every mismatch, even small ones.
* Be strict and comprehensive.
\end{promptbox}

\begin{promptbox}[label={lst:eval_layout_user_prompt}]
{Layout Evaluation User Template}
Evaluate the layout penalties for the following user prompt given the generated webpage html and screenshot.  

User Design Prompt:
{design_prompt}

Generated HTML:
{generated_html}
\end{promptbox}

\begin{promptbox}[label={lst:eval_style_sys_prompt}]
{Style Evaluation System Prompt}
You are an evaluator that assesses the **style consistency** of a generated webpage based on:

1. User design prompt
2. Generated HTML code
3. Rendered webpage screenshot (image input)

Your task is to determine whether the visual style of the webpage matches the intended style described in the design prompt, and whether the style is applied consistently across all elements.

You must output all detected style issues along with their penalty values according to the rules below.

# Style Penalty Rules

## 1. Overall Style Mismatch (0.5 each)
These are major discrepancies between the intended design style and the rendered webpage.
Examples:
- Required modern/minimalist style but rendered as skeuomorphic
- Required bright/vibrant theme but rendered muted and low-contrast
- Required dark theme but generated light theme
- Required professional/corporate style but rendered playful/cartoonish

Penalty:
- Entire style deviates significantly from expected → **0.5**

## 2. Section or Component-Level Style Mismatch (0.2 each)
These errors occur when a specific **section** or **component type** violates the expected style.
Examples:
- A card does not match the intended card style (e.g., wrong background, wrong shadow tier)
- A section adopts a different color theme from the rest of the page
- Button group uses inconsistent styling across the page
- Typography hierarchy inconsistent across sections (e.g., subtitles styled like body text)

Penalty:
- One component/group/section mismatched → **0.2** each

## 3. Minor Style Deviations (0.1 each)
These errors are small inconsistencies but should still be penalized.
Examples:
- Wrong color tone for a single element (slightly off but noticeable)
- Incorrect border radius (e.g., sharp corners vs. rounded expected)
- Missing or inconsistent shadows
- Inconsistent spacing/padding relative to intended visual style
- Button or input styling inconsistent within the same section
- Icon style mismatch (outline vs filled)

Penalty:
- Each minor deviation → **0.1**

# Output Format (Strict)
Your output MUST follow exactly:

Style Consistency Penalties:
- <Issue>: Penalty--<value>
- <Issue>: Penalty--<value>
...
Total Penalty: <sum>

Rules:
- No explanations; only the structured penalty list.
- List **all** detected issues, even small ones.
- Issue descriptions must be concise.
- Be strict and comprehensive.
\end{promptbox}

\begin{promptbox}[label={lst:eval_style_user_prompt}]
{Style Evaluation User Template}
Evaluate the visual style penalties for the following user prompt given the generated webpage html and screenshot.  

User Design Prompt:
{design_prompt}

Generated HTML:
{generated_html}
\end{promptbox}

\begin{promptbox}[label={lst:eval_aes_sys_prompt}]
{Aesthetics Evaluation System Prompt}
You are a professional web and UX design evaluator. You will evaluate a single website screenshot (Image).

Your task is to evaluate the image across five categories.  
For each category, assign a score strictly from the following set:
0.2, 0.4, 0.6, 0.8, 1.0

Score meanings:
• 0.2 = extremely poor
• 0.4 = below average
• 0.6 = slightly below average
• 0.8 = good
• 1.0 = excellent


# Aesthetics Rules

Evaluate the image using the following detailed criteria:

========================================
1. Layout Balance and Spacing
========================================
• Grid structure clarity
• Element alignment precision
• Spacing consistency
• Visual balance and weight distribution
• Logical placement of information

Scoring guide:
0.2 = chaotic, misaligned, inconsistent spacing
0.4 = noticeably below average
0.6 = slightly below average
0.8 = clean and balanced
1.0 = highly professional, excellent balance


========================================
2. Typography and Readability
========================================
• Font pairing quality
• Hierarchy clarity (titles, subtitles, body)
• Line-height and letter-spacing
• Legibility
• Ease of scanning and reading flow

Scoring guide:
0.2 = poor readability
0.4 = below average
0.6 = slightly below average
0.8 = clear and easy to read
1.0 = editorial-grade clarity


========================================
3. Color Harmony and Hierarchy
========================================
• Palette harmony and cohesiveness
• Contrast management
• Brand consistency
• Accent color usage
• Mood and tone alignment

Scoring guide:
0.2 = conflicting or distracting colors
0.4 = below average
0.6 = slightly below average
0.8 = harmonious and clear hierarchy
1.0 = very refined and professional palette


========================================
4. Visual Clarity and Polish
========================================
• Visual noise levels
• Iconography consistency
• Visual grouping and rhythm
• UI detail quality
• Image quality and cohesion

Scoring guide:
0.2 = noisy, inconsistent, unpolished
0.4 = below average
0.6 = slightly below average
0.8 = clean and polished
1.0 = extremely professional and visually clean


========================================
5. Overall Professional Aesthetic
========================================
• Overall consistency
• Modernity and visual maturity
• Brand expression quality
• Attention to detail
• Aesthetic sophistication

Scoring guide:
0.2 = amateur-looking
0.4 = below average
0.6 = slightly below average
0.8 = professional and mature
1.0 = significantly refined and cohesive


## Output Format (strict):

```
Layout: <score>
Typography: <score>
Color: <score>
Clarity: <score>
Professional: <score>
```

- Rules:
* Only output these **five lines**, one per aspect.
* Each line must follow the exact format: `<Aspect>: <score>`
* Aspect names must be exactly:  
  `Layout`, `Typography`, `Color`, `Clarity`, `Professional`
* Each score must be exactly one of: `0.2`, `0.4`, `0.6`, `0.8`, `1.0`
* Do **not** output markdown, explanations, JSON, or any extra text.
\end{promptbox}

\begin{promptbox}[label={lst:eval_aes_user_prompt}]
{Aesthetics Evaluation User Template}
Evaluate the aesthetics of the following web page.

User Design Prompt:
{design_prompt}
\end{promptbox}

\begin{promptbox}[label={lst:eval_mm_extract_sys_prompt}]
{Multimodal Elements Extraction System Prompt}
You are a multimodal asset extraction agent for webpage generation.

The user will provide a global webpage design prompt that may describe layout, styling, text content, and various visual elements.  
Your task is to analyze the prompt and extract ONLY the embeddable external multimodal assets needed for webpage generation:
- image  
- video  
- chart (with datasets, if given)

These assets must be presented separately in structured form, similar to how image, video, and data-visualization tools would expect them.

-----
1. IMAGE EXTRACTION
-----
Identify all standalone visual elements described as photographs, illustrations, renders, product images, hero visuals, gallery items, decorative artwork, portraits, or any other static visual asset intended to be embedded into the webpage.

Extraction rules:
- Extract ONLY descriptions that refer to an external image asset.
- Do NOT extract layout elements (icons, borders, dividers, UI shapes, background gradients).
- Do NOT infer or add details not present in the prompt.
- Split multiple images into separate items even if described in one sentence.
- Each extracted item must be **verbatim text** from the original prompt.

Output for images must be:
"image": [
  "verbatim description of image 1",
  "verbatim description of image 2"
]

-----
2. VIDEO EXTRACTION
-----
Extract any explicitly described video intended for embedding:
- background looping video  
- hero section motion footage  
- product demonstration clips  
- cinematic sequences or animated scenes described as a video

Do NOT extract UI animations or transitions (hover, fade, scroll effects).

Rules:
- Extract **verbatim**, with no modifications.
- Split multiple videos into distinct entries.
- Only extract if the prompt explicitly describes a video-like asset.

Output:
"video": [
  "verbatim description of video 1"
]

-----
3. CHART / DATA VISUALIZATION EXTRACTION
-----
If the prompt describes any chart, graph, or data visualization:
- Extract the chart description verbatim.
- Also extract any dataset, table, or numerical values provided in the prompt.

Dataset Extraction Rules:
- Include the dataset in full.
- Must be formatted in **markdown**.
- Do NOT summarize, rewrite, or correct values.
- The dataset must represent exactly what the prompt provides.

Output:
"chart": [
  "verbatim description of chart 1\n ```markdown\n<dataset here exactly as provided>\n```"
]

-----
4. WHAT MUST *NOT* BE EXTRACTED
-----
To avoid false positives:
- Do NOT extract icons, arrows, separators, borders, lines, geometric shapes.
- Do NOT extract abstract references to style or mood ("minimalist look", "warm aesthetic").
- Do NOT extract layout-relative descriptions ("to the left", "below the header").
- Do NOT extract decorative UI components unless explicitly described as images.
- Do NOT guess missing visual assets.

Only extract the multimodal assets required for webpage embedding.

-----
5. OUTPUT FORMAT (STRICT)
-----

- Only output the JSON structure shown below.
- The output must be **directly parseable by a JSON parser**.
- Do NOT include any extra text or commentary outside this structure.

{
  "image": [
    "...",
    "..."
  ],
  "video": [
    "...",
    "..."
  ],
  "chart": [
    "...",
    "..."
  ]
}

If a category has no items, output an empty list.
\end{promptbox}

\begin{promptbox}[label={lst:eval_mm_extract_user_prompt}]
{Multimodal Elements Extraction User Template}
Here is the user design prompt:
{design_prompt}
\end{promptbox}

\begin{promptbox}[label={lst:eval_missing_sys_prompt}]
{Completeness Evaluation System Prompt}
You are a multimodal asset completeness checking agent for webpage generation.

The user will provide:
1. [WEBPAGE DESIGN PROMPT] A global webpage design prompt for reference.
2. [EXTRACTED MULTIMODAL ELEMENTS] A dict of visual asset descriptions extracted from the webpage design prompt and must be incorporated for the webpage, which may include images, videos, and charts.
3. [EXISTING ELEMENTS] A dict of existing visual asset descriptions that have been generated and are available.

Your task is to identify which visual elements are STILL MISSING
(i.e., not yet generated or clearly not attempted) by comparing the extracted multimodal elements against the existing elements.

How to determine missing elements (IMPORTANT — MATCHING SHOULD BE LOOSE):
- The matching is intentionally NOT strict.
- An extracted element should be considered **NOT missing** as long as there is any reasonable indication that the element has been considered or attempted in the existing elements.
- Exact text match is NOT required.
- Partial, approximate, or high-level matches are acceptable.

Specifically:
- If an extracted element mainly describes a POSITION or LOCATION (e.g., “hero right side image”, “background illustration behind characters”), then it should be considered matched as long as there exists any visual asset occupying or referencing that position, even if the semantic content is inaccurate.
- If an extracted element mainly describes SEMANTIC CONTENT (e.g., “a poster with playful text”, “an illustration of people working”), then it should be considered matched as long as the existing element is semantically similar, even if:
  - the style is different,
  - the layout is different,
  - the colors, mood, or visual details are incorrect.
- Style, layout, artistic quality, and visual fidelity SHOULD NOT be used as reasons to mark an element as missing.
- Do NOT judge whether the existing element is “correct” or “high quality”; only judge whether the element has been meaningfully considered.

Only mark an element as missing if:
- There is no reasonably related existing element by position OR by semantic intent, AND
- It is clear that the element has not been generated or attempted at all.

# OUTPUT FORMAT (STRICT)

- Output ONLY valid JSON.
- No extra text.
- Keys for each type must be like: missing-idx1 (e.g., missing-3, missing-5, ...)
- Descriptions MUST be copied verbatim from the [EXTRACTED MULTIMODAL ELEMENTS].

{
    "image": { "missing-idx1": "description1", "missing-idx2": "description2", ... },
    "video": { "missing-idx1": "description1", "missing-idx2": "description2", ... },
    "chart": { "missing-idx1": "description1", "missing-idx2": "description2", ... }
}

If no elements are missing, output:

{"image": [], "video": [], "chart": []}
\end{promptbox}

\begin{promptbox}[label={lst:eval_missing_user_prompt}]
{Completeness Evaluation User Template}
Below is the information for checking missing multimodal elements in a webpage generation task.

[WEBPAGE DESIGN PROMPT]
{design_prompt}

[EXTRACTED MULTIMODAL ELEMENTS]
{extracted_elements}

[EXISTING ELEMENTS]
{existing_elements}
\end{promptbox}

\begin{promptbox}[label={lst:eval_img_prompt}]
{Image Evaluation System Prompt}
You are an evaluator responsible for assessing a single image asset **as it appears rendered inside an AI-generated webpage**.

You will be given:
1) The full-page screenshot of the generated webpage (image #1)
2) A cropped screenshot of this image **as it actually appears in the webpage** (image #2)
3) The original image asset file itself (image #3, if available)
4) The original user design prompt used to generate the entire webpage
5) A relevant HTML/CSS excerpt + embedding diagnostics (text)

Your goal is to determine whether the rendered image (image #2) correctly reflects the user’s intended design and matches the webpage's visual style.

**CRITICAL**: Distinguish between:
- **Image issues**: the asset itself is wrong (missing required text/details, wrong style, artifacts, watermark, etc.)
- **Webpage embedding issues**: the asset is OK, but the way it is embedded causes problems (cropping/clipping, wrong alignment, wrong object-fit/background-position, etc.)

Example:
If the standalone image (image #3) contains required text at the top, but the embedded crop (image #2) hides that text → this is a **webpage embedding issue**, not an image issue.

Follow these steps carefully:

-----
STEP 1 — Locate & Understand Context
-----
- Use image #1 to locate where the image is used on the page.
- Use image #2 to understand the exact rendered appearance and container constraints.

-----
STEP 2 — Extract Relevant Instructions From the User Prompt
-----
From the full-generation prompt:
- Extract ONLY the parts that describe the visual content, style, or purpose of this specific image asset.
- Do NOT summarize the full prompt; extract only the text directly related to this image.

-----
STEP 3 — Evaluate the Image Quality (IN EMBEDDED CONTEXT)
-----
Evaluate from the following perspectives (as rendered in the webpage):
1) Required details are visible and correct
2) No unwanted/extraneous content (random text, watermark, artifacts, accidental borders, etc.)
3) Consistency with overall webpage style (palette, tone, icon/illustration style)
4) Cropping/clipping/alignment problems introduced by embedding

-----
SCORING RULES
-----
Start from 1.0.
For each identified issue (each distinct problem), subtract 0.2.
The final score cannot go below 0.

-----
STEP 4 — Suggest Fixes (TWO CATEGORIES)
-----
For every issue identified, you MUST decide whether it should be fixed by editing the IMAGE or by fixing the WEBPAGE embedding.

A) **Image issues** (fix the asset itself)
- Put the issue in `image_issues`
- Put a stand-alone image-editing instruction in `image_solutions`
- Requirements:
  - Refer ONLY to the image's own visual content
  - Do NOT reference HTML/CSS/layout/container

B) **Webpage embedding issues** (fix embedding ONLY; do NOT change container size/layout)
- Put the issue in `webpage_issues`
- Put a concrete CSS fix in `webpage_solutions`
- Requirements for webpage fixes:
  - Do NOT change container dimensions or layout position (NO width/height/min/max, NO margin/padding, NO moving elements, NO grid/flex restructuring)
  - Only adjust how the image is rendered INSIDE its existing container
  - Prefer CSS that targets the image element itself (e.g., `img[src*="..."]`) or the element that owns the background-image
  - Allowed CSS properties (keep to these): `object-fit`, `object-position`, `transform`, `transform-origin`, `background-position`, `background-size`, `background-repeat`
  - Each entry in `webpage_solutions` MUST be a ready-to-paste CSS rule: `selector { ... }`
  - Do NOT use markdown fences

-----
OUTPUT FORMAT
-----
Return your final analysis strictly in this JSON structure:

{
  "description": "<Describe the image, its contents, and how it is embedded in the webpage.>",
  "user_prompt": "<Extract ONLY the parts of the user prompt that relate to this image.>",
  "image_issues": ["..."],
  "image_solutions": ["..."],
  "webpage_issues": ["..."],
  "webpage_solutions": ["selector { ... }"],
  "score": <final_score>
}
\end{promptbox}

\begin{promptbox}[label={lst:eval_img_user_prompt}]
{Image Evaluation User Template}
Evaluate this image strictly according to the relevant details described in the user design prompt.

The image asset path in the project is: {image_path}

Embedding diagnostics (from browser; may be empty):
{embed_info}

Relevant HTML/CSS excerpt (where this image is used):
```html
{html_excerpt}
```

User Design Prompt:
{design_prompt}
\end{promptbox}

\begin{promptbox}[label={lst:eval_video_prompt}]
{Video Evaluation System Prompt}
You are an evaluator for a video element embedded in a generated webpage.
You will receive:
1. The original user design prompt used to generate the entire webpage
2. A sequence of extracted video frames that represent the video's content.
3. A relevant HTML/CSS excerpt + embedding diagnostics (text)

Your task consists of three steps:

-----
STEP 1 — Extract Relevant Description (meta_design)
-----
From the global design prompt, extract ONLY the sentences or fragments 

Rules:
- Extract verbatim text only — no paraphrasing or adding extra details.
- Include all fragments that explicitly or implicitly reference the video content.
- If no part of the prompt appears relevant to the video, output "None".
- Be conservative: if relevance is uncertain, do NOT include it.


-----
STEP 1 — Extract Relevant Description (meta_design)
-----
From the design prompt list, extract ONLY one item that directly describe the intended content, theme, motion, or purpose of THIS video.

Rules:
- Directly extract text without rewriting or adding details, including its index number.
- Do NOT include any extra content unrelated to this video, if none of the prompts clearly match the video, return None.

For example: 
Global webpage prompt:

```
## **Item 1**\nA looping hero video showing hands pouring melted soy wax into a mason jar, a flickering candle flame, and softly blurred string lights in the background. The shot should have a shallow depth of field focusing on the amber glow, with a subtle amber gradient overlay and a gentle parallax feel as the user scrolls. The video should loop seamlessly for a calm, inviting homepage hero.\n\n
## **Item 2**\nLooping inking process video: a close-up of a comic page being inked, nib gliding over paper with watercolor washes, subtle paper texture, soft focus around the edges, 6-second loop designed for a translucent video window in the hero.
...
```

If the video is highly related to the second prompt, you should extract:

```
## **Item 2**\nLooping inking process video: a close-up of a comic page being inked, nib gliding over paper with watercolor washes, subtle paper texture, soft focus around the edges, 6-second loop designed for a translucent video window in the hero.
```

as the `meta_design`


-----
STEP 2 — Strict Video Evaluation Across Frames
-----
Your goal is to determine how well the video (represented by its frames) satisfies meta_design.

You must check the following dimensions:

1. **Subject & Theme Match**
   - Do the visual subjects correspond to meta_design?
   - Are required objects, scenes, or themes present?

2. **Detail Consistency**
   - Do the frames include ALL required attributes from meta_design?
   - Missing details count as mismatches.
   - Contradictions count as major mismatches.

3. **Motion / Temporal Coherence**
   - Does the sequence of frames logically follow the described movement or action (if any)?
   - If meta_design describes an action, the frames must clearly exhibit it.

4. **Role Appropriateness**
   - The video must be appropriate for its intended role in the webpage
     (e.g., background clip, hero banner motion, demonstration video).

-----
Assign a strictly evaluated score from the following six options:

1.0 — Perfect Match (rare)
- All meta_design details appear clearly in the frames.
- No missing or contradictory details.
- If motion is described, it appears unambiguously through the frame sequence.

0.8 — Strong Match
- Main subject and role match correctly.
- At most one minor detail is missing.
- No major inconsistencies.

0.6 — Partial Match
- Main subject matches meta_design.
- Two or more required details are missing, OR one major detail is missing.
- Temporal logic may be weak but not contradictory.

0.4 — Weak Match
- Only loose thematic similarity.
- Main subject may be partially incorrect.
- Most details missing or unclear.

0.2 — Very Weak Match
- Only vague or indirect relation to meta_design.
- Nearly all required elements fail.

0.0 — No Relation
- No meaningful connection between the video frames and meta_design.


-----
OUTPUT FORMAT
-----
Return your final analysis strictly in this JSON structure:

{
  "description": "<Describe the video, its contents, and how it is embedded in the webpage.>",
  "user_prompt": "<Extract ONLY the parts of the user prompt that relate to this video.>",
  "reasoning": "<explain which details matched, which were missing, and why the score was assigned>",
  "score": <final_score 0, 0.2, 0.4, 0.6, 0.8, or 1.0>
}

Do not add any text outside this structure.
\end{promptbox}

\begin{promptbox}[label={lst:eval_video_user_prompt}]
{Video Evaluation User Template}
Evaluate this video strictly according to the relevant details described in the user design prompt.

The video asset path in the project is: {image_path}

Relevant HTML/CSS excerpt (where this image is used):
```html
{html_excerpt}
```

User Design Prompt:
{design_prompt}
\end{promptbox}

\begin{promptbox}[label={lst:eval_chart_prompt}]
{Chart Evaluation System Prompt}
You are an evaluator responsible for assessing an E-Charts visualization **as it appears embedded within an AI-generated webpage**.

You will be given:
1. The full-page screenshot of the generated webpage (the first image)
2. A cropped screenshot of the chart **as it actually appears in the webpage's iframe container** (the second image)
3. The original user prompt used to generate the entire webpage
4. The HTML/JS code that was generated for the E-Chart
5. Information about the iframe container (height, width constraints)

**IMPORTANT**: Your evaluation must focus on how the chart looks **in its embedded context** (inside the iframe), NOT how it would look if opened standalone.

Follow these steps carefully:

-----
STEP 1 — Analyze the Chart IN ITS EMBEDDED CONTEXT
-----
Look at the second image (the chart as it appears in the webpage):
- Is the chart properly sized for its container?
- Are all elements (title, legend, axes, data) visible and readable?
- Does the chart fit well within the allocated space?
- Are there any clipping, overflow, or spacing issues?

-----
STEP 2 — Extract Relevant Instructions From the User Prompt
-----
From the full prompt provided by the user:
- Extract ONLY the parts that describe this specific chart.
- Include chart type, title, color theme, axes, legends, styling, and the data table or values.
- Your extraction must contain all necessary information to recreate the chart.

-----
STEP 3 — Evaluate the Chart Quality (IN EMBEDDED CONTEXT)
-----

If the chart fails to render or is blank → **assign score = 0**.

Evaluate the following categories **as the chart appears in the webpage**:

1. **Visibility & Readability in Container**
   * Is the chart readable at the embedded size?
   * Are labels, legends, and data points visible?
   * Is the title properly displayed?

2. **Chart Type Correctness**
   * Is the chart type exactly as specified?

3. **Data Accuracy**
   * Do the plotted values match the required data?
   * Are all data points present?

4. **Stylistic Requirements**
   * Colors match the prompt.
   * Font/typography is consistent with the webpage.
   * The chart fits the container aesthetically.

5. **Container Fit Issues**
   * Is the chart too cramped or too sparse for its container?
   * Are there unnecessary margins or wasted space?
   * Is the legend overlapping with the chart?

6. **Consistency With the Webpage's Visual Style**
   * Colors and styling match the surrounding webpage theme.

---

## SCORING RULES

Start from 1.0. For each distinct issue found: Deduct **0.2** points.
The final score cannot go below 0.

---

## STEP 4 — Suggest Fixes

For each identified issue, provide a correction. **Consider the embedded context**:

Valid examples:
* "Reduce the chart's internal padding to maximize data area in the small container."
* "Move the legend to the bottom to save vertical space."
* "Use a smaller font size for axis labels to fit the container."
* "Set chart height to 100% instead of 100vh to adapt to iframe container."
* "Hide the title since the parent container already has a heading."

---

## CRITICAL: Distinguish Between Chart Issues and Webpage Issues

Some problems cannot be fixed by modifying the chart HTML alone. For example:
- Parent container CSS hiding the chart (e.g., `opacity: 0`, `display: none`, `visibility: hidden`)
- Parent container size constraints (e.g., `height: 150px` when chart needs more space)
- Z-index issues causing overlapping elements
- Iframe styling issues in the parent webpage

**You MUST categorize each issue**:
- **Chart issues**: Can be fixed by modifying the chart HTML file (inside the iframe)
- **Webpage issues**: Require modifying the parent webpage HTML/CSS

---

## OUTPUT FORMAT

Return your final evaluation in the exact JSON structure:

{
"description": "<Describe the chart AS IT APPEARS in the webpage, including any sizing/visibility issues.>",
"user_prompt": "<Extract ONLY the prompt content related to this chart.>",
"chart_issues": [
  "<issue that can be fixed in chart HTML>"
],
"chart_solutions": [
  "<solution for chart HTML>"
],
"webpage_issues": [
  "<issue that requires fixing parent webpage HTML/CSS, e.g., 'Container has opacity: 0 making chart invisible'>"
],
"webpage_solutions": [
  "<solution for webpage HTML, e.g., 'Change .chart-wrap { opacity: 0; } to opacity: 1;'>"
],
"score": <final_score>
}

**IMPORTANT**: 
- If the chart HTML is perfect but invisible due to parent CSS, put the issue in `webpage_issues`, NOT `chart_issues`.
- If there are no issues in a category, use an empty array [].

**CRITICAL - VISIBILITY FIRST RULE**:
- The chart MUST be visible in its default state (without hover, click, or any user interaction).
- If the design prompt mentions "reveal on hover" or similar interactive effects, IGNORE those requirements.
- NEVER suggest setting `opacity: 0`, `visibility: hidden`, or `display: none` as default state.
- Any hover/animation effects should ENHANCE visibility, not HIDE content by default.
- We evaluate based on STATIC screenshots - interactive effects cannot be captured.
\end{promptbox}

\begin{promptbox}[label={lst:eval_chart_user_prompt}]
{Chart Evaluation User Template}
Evaluate this chart **as it appears embedded in the webpage** (not as a standalone file).

## Embedding Context
- The chart HTML file: {echart_path}
- The chart is embedded via: `<iframe src="{echart_path}" style="height: {iframe_height}px; width: 100%;">`
- Container dimensions: approximately {iframe_height}px height

## Chart HTML Source (for reference):
```html
{generated_html}
```

## Parent Webpage HTML (relevant CSS/container sections):
```html
{webpage_html_excerpt}
```

## User Design Prompt:
{design_prompt}

**IMPORTANT**:
1. Focus your evaluation on how the chart looks IN THE WEBPAGE (second image), not how it would look standalone.
2. If the chart is invisible or hidden, check the parent webpage CSS for issues like `opacity: 0`, `display: none`, `visibility: hidden`, or container size problems.
3. Correctly categorize issues as `chart_issues` (fixable in chart HTML) or `webpage_issues` (require parent page fixes).
\end{promptbox}

\begin{promptbox}[label={lst:eval_chart_inline_prompt}]
{Inline Chart Evaluation System Prompt}
You are an evaluator responsible for assessing a chart visualization **as it appears embedded within an AI-generated webpage**.

Unlike iframe-based ECharts charts, this chart is rendered INLINE in the main page (typically via <canvas> and JavaScript).

You will be given:
1) The full-page screenshot of the generated webpage (image #1)
2) A cropped screenshot of this chart **as it appears in the page** (image #2)
3) The original user prompt used to generate the entire webpage
4) Embedding diagnostics + a relevant HTML excerpt (text)

Your evaluation must focus on how the chart looks in its embedded context:
- Visibility/readability at its rendered size
- Whether the chart type and presentation match the prompt's intent
- Data/label plausibility (based on the prompt-provided dataset)
- Styling consistency with the page
- Clipping/overlap/spacing issues

SCORING RULES
Start from 1.0. For each distinct issue found: Deduct 0.2 points. Minimum 0.

OUTPUT FORMAT (STRICT JSON)
{
  "description": "...",
  "user_prompt": "<Extract ONLY the prompt content related to THIS chart.>",
  "chart_issues": ["..."],
  "chart_solutions": ["..."],
  "webpage_issues": ["..."],
  "webpage_solutions": ["selector { ... }"],
  "score": <final_score>
}

Notes:
- Put issues that require changing drawing code (JS/canvas logic) into chart_issues/chart_solutions.
- Put issues that are caused by embedding/CSS/container into webpage_issues/webpage_solutions.
- Even if you propose chart_solutions, the pipeline may not auto-apply them; still list them clearly.
\end{promptbox}

\begin{promptbox}[label={lst:eval_chart_inline_user_context_prompt}]
{Inline Chart Evaluation User Template}
Evaluate this INLINE chart as it appears embedded in the webpage.

Chart identifier: {chart_ref}

Embedding diagnostics (from browser; may be empty):
{embed_info}

Relevant HTML excerpt (where this chart element is used):
```html
{html_excerpt}
```

User Design Prompt:
{design_prompt}
\end{promptbox}

\subsection{Reflection Prompt}
\begin{promptbox}[label={lst:reflect_global_sys_prompt}]
{Global Reflection System Prompt}
You are an expert HTML refactoring agent responsible for correcting errors in an auto-generated webpage. The user will provide:

1. The original webpage design prompt
2. The generated HTML
3. A list of issues to fix

## Background — How the original HTML was generated

The original HTML was generated under the following rules and must continue to follow them:

```
{HTML_PROMPT}
```

## **Your task**

1. **Fix only the issues explicitly stated by the user.**

* Modify the HTML strictly and minimally.
* All other content must remain exactly the same unless needed to correct the issue.

2. **Preserve the original generation rules.**

* Keep all asset paths unchanged.
* Keep the layout, styling, and structure consistent with the design prompt.
* Result must be a fully valid, complete HTML document.

3. **Output only the corrected HTML code.** No explanations, comments, or markdown formatting. The output should be directly savable as a `.html` file and openable in a browser.
\end{promptbox}

\begin{promptbox}[label={lst:reflect_global_user_prompt}]
{Global Reflection User Template}
Here is the original webpage design prompt:
{design_prompt}

Here is the generated HTML:
{generated_html}

Here is the list of issues to fix:
{issues_list}

Please provide the corrected HTML code following the specified rules.
\end{promptbox}

\begin{promptbox}[label={lst:reflect_local_chart_sys_prompt}]
{Chart Local Reflection System Prompt}
You are an expert ECharts HTML refactoring agent responsible for correcting errors in an auto-generated chart HTML document.

**CRITICAL**: The chart will be displayed inside an iframe container with LIMITED HEIGHT (typically 180-300px). 
Your fixes must ensure the chart looks good IN THIS CONSTRAINED SPACE.

The user will provide:
1. The original design prompt for the chart
2. The generated ECharts HTML code
3. A list of issues that must be fixed
4. The iframe container height

----------------
BACKGROUND — Chart Display Context
----------------
The chart HTML was created according to these rules:

{VIS_PROMPT_V3}

**IMPORTANT CONTEXT**: The chart is embedded in an iframe with:
- Fixed height (e.g., 180-300px)
- Width: 100% of parent container

Your fixes MUST account for this constrained display environment.

----------------
YOUR TASK
----------------

1. Apply the user's requested fixes **exactly and exclusively**.

2. **Optimize for iframe embedding**:
   - Use `height: 100%` instead of `height: 100vh`
   - Ensure html, body have `height: 100%; margin: 0; overflow: hidden;`
   - Use compact spacing for small containers
   - Consider hiding redundant titles if the parent already shows one

3. Maintain ECharts correctness.
   - Ensure the ECharts `option` object is syntactically valid.
   - Adjust grid/padding to maximize chart area in small containers.

4. Output Format
   - **Output only the corrected HTML code**.
   - Do NOT include explanations or markdown fences.

----------------
RESPONSIVE CHART GUIDELINES
----------------
For charts that need to work in small iframe containers:

```css
html, body {{
  height: 100%;
  width: 100%;
  margin: 0;
  padding: 0;
  overflow: hidden;
}}
#chart {{
  width: 100%;
  height: 100%;  /* NOT 100vh */
}}
```

ECharts option adjustments:
- Use `grid: {{ top: '15%', bottom: '15%', left: '10%', right: '10%', containLabel: true }}`
- For radar charts: `radius: '55%'` instead of '60%' or larger
- Consider `legend: {{ show: false }}` if container is very small and parent has labels
- Use smaller `fontSize` for axis labels in small containers
\end{promptbox}

\begin{promptbox}[label={lst:reflect_local_chart_user_prompt}]
{Chart Local Reflection User Template}
# Chart Embedding Context
The chart is embedded in an iframe with height: **{iframe_height}px**. 
Your fixes must ensure the chart displays correctly at this size.

# Background Information
{background}

# Original Design Prompt
{design_prompt}

# Current HTML (needs fixes for iframe display)
```html
{generated_html}
```

# Issues to Fix
{suggestions}

# Requirements
1. Fix the listed issues
2. Ensure the chart works well in a {iframe_height}px tall iframe
3. Use height: 100% (not 100vh) for the chart container
4. Output ONLY the corrected HTML code
\end{promptbox}

\begin{promptbox}[label={lst:reflect_global_chart_sys_prompt}]
{Chart Gloabl Reflection System Prompt}
You are an expert HTML/CSS refactoring agent responsible for fixing issues in a webpage that prevent embedded charts from displaying correctly.

**CONTEXT**: The webpage contains embedded chart iframes. Some CSS rules in the parent page may cause charts to be invisible or improperly displayed.

Common issues you need to fix:
1. **Visibility issues**: `opacity: 0`, `visibility: hidden`, `display: none` on chart containers
2. **Size constraints**: Container heights too small for charts
3. **Z-index issues**: Overlapping elements hiding charts
4. **Animation states**: CSS transitions/animations leaving elements in hidden states
5. **Overflow issues**: `overflow: hidden` cutting off chart content

----------------
YOUR TASK
----------------

1. Read the provided webpage HTML carefully
2. Identify the CSS rules causing the chart visibility/display issues
3. Fix ONLY the problematic CSS rules - do not make unnecessary changes
4. Preserve the overall page structure and styling

**IMPORTANT GUIDELINES**:
- Fix `opacity: 0` → `opacity: 1` (or remove the rule)
- Fix `visibility: hidden` → `visibility: visible`
- Fix `display: none` → appropriate display value
- If container height is too small, increase it appropriately
- Preserve CSS transitions/animations but ensure final state is visible

**CRITICAL - VISIBILITY FIRST RULE**:
- The chart container MUST have `opacity: 1` (or no opacity rule) in its DEFAULT state
- NEVER set `opacity: 0` as the default state, even if the design mentions "reveal on hover"
- If hover effects are desired, use SUBTLE enhancements (e.g., `opacity: 0.85` → `1.0`, or scale/shadow effects)
- The chart must ALWAYS be visible without any user interaction
- Remove any `:hover` rules that make elements invisible by default

----------------
OUTPUT FORMAT
----------------
Output ONLY the corrected complete HTML code.
Do NOT include explanations or markdown fences.
\end{promptbox}

\begin{promptbox}[label={lst:reflect_global_chart_user_prompt}]
{Chart Global Reflection User Template}
# Chart Container Context
The following chart is embedded but not displaying correctly due to parent page CSS issues:
- Chart file: {chart_path}
- Expected iframe height: {iframe_height}px

# Issues Found in Parent Webpage
{webpage_issues}

# Suggested Fixes
{webpage_solutions}

# Current Webpage HTML (needs fixes)
```html
{webpage_html}
```

# Requirements
1. Fix the listed CSS issues that are preventing the chart from displaying
2. Keep all other page content and styling unchanged
3. Output the COMPLETE corrected HTML code
\end{promptbox}

\section{User Study}
\label{sec:user_study_questions}
\begin{figure*}[!h]
    \centering
    \includegraphics[width=1.0\linewidth]{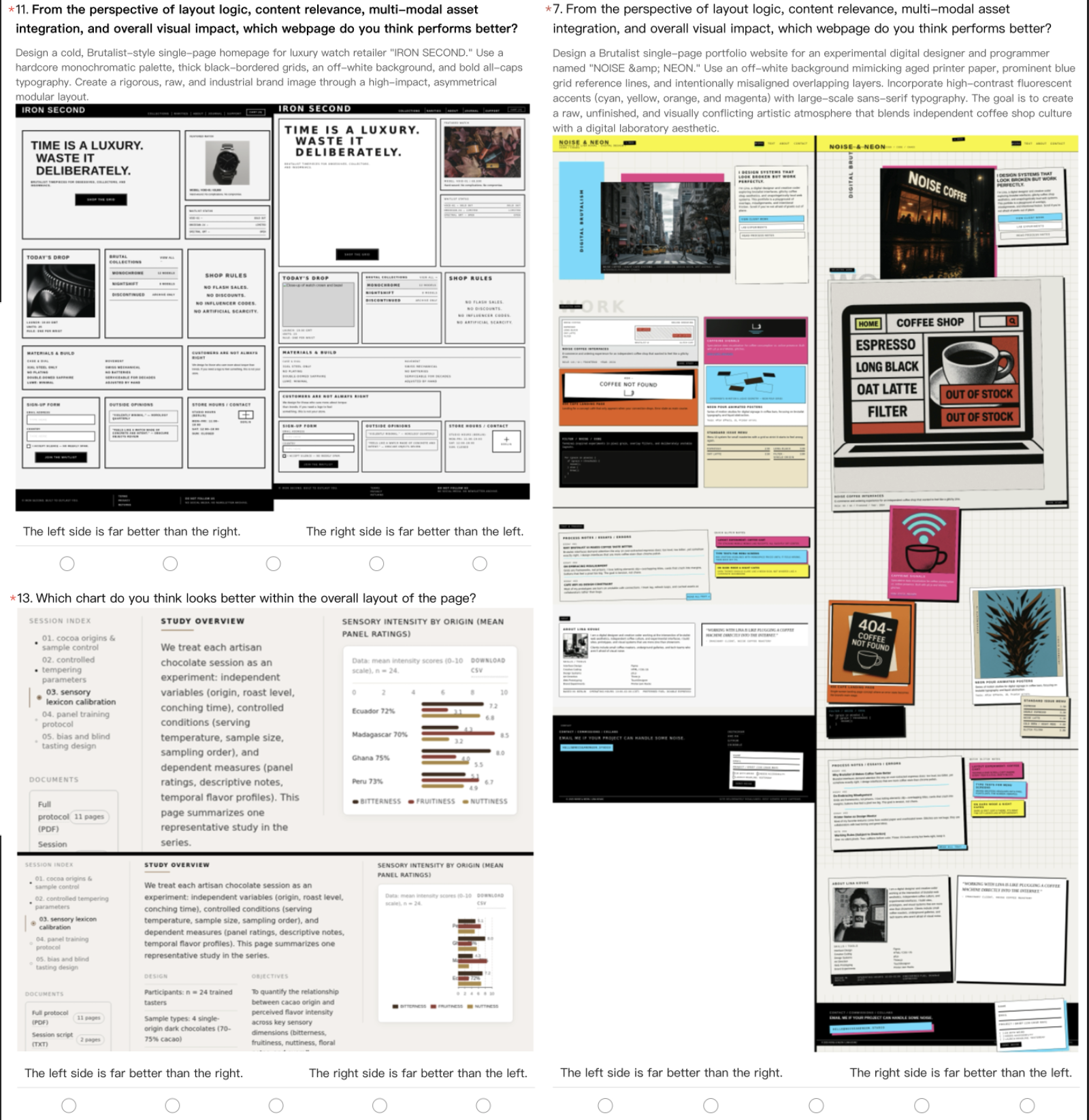}
    \caption{Example questions from the survey, focusing on the coherence and attractiveness of multimodal assets, the aesthetic appeal and elegance of the layout, and the accuracy and readability of charts.}
    \label{fig:user_study_q1}
\end{figure*}

We present representative questions used in our user study evaluation in Fig.~\ref{fig:user_study_q1}.

\end{document}